\pdfoutput=1

\documentclass[11pt]{article}

\usepackage[final]{acl}

\usepackage{times}
\usepackage{latexsym}
\usepackage{listings}
\usepackage{fancyvrb}
\usepackage[T1]{fontenc}

\usepackage[utf8]{inputenc}
\usepackage{academicons}
\usepackage{fontawesome5}
\usepackage{hyperref}

\usepackage{microtype}

\usepackage{inconsolata}
\usepackage{alltt}

\usepackage{subfig}
\usepackage{booktabs}
\usepackage{graphicx}
\usepackage{multirow}
\usepackage{tabularx}
\usepackage{array} 
\usepackage{amsmath}
\usepackage{amssymb}
\usepackage{microtype}
\usepackage{inconsolata}
\usepackage{longtable}
\usepackage{makecell}   

\usepackage{xcolor}

\title{Comparing human and LLM politeness strategies in free production}

\author{Haoran Zhao \\
  Department of Linguistics \\
  University of Washington \\
  \texttt{hjzhao@uw.edu} \\\And
  Robert D. Hawkins \\
  Department of Linguistics \\
  Stanford University \\
  \texttt{rdhawkins@stanford.edu} \\}

\begin{document}
\maketitle
\begin{abstract}
Polite speech poses a fundamental alignment challenge for large language models (LLMs). 
Humans deploy a rich repertoire of linguistic strategies to balance informational and social goals – from positive approaches that build rapport (compliments, expressions of interest) to negative strategies that minimize imposition (hedging, indirectness).
We investigate whether LLMs employ a similarly context-sensitive repertoire by comparing human and LLM responses to English-language scenarios in both constrained and open-ended production tasks.
We find that larger models ($\ge$70B parameters) successfully replicate key effects from the computational pragmatics literature, and human evaluators prefer LLM-generated responses in open-ended contexts. 
However, further linguistic analyses reveal that models disproportionately rely on negative politeness strategies to create distance even in positive contexts, potentially leading to misinterpretations. 
While LLMs thus demonstrate an impressive command of politeness strategies, these systematic differences provide important groundwork for making intentional choices about pragmatic behavior in human-AI communication.

\end{abstract}

\section{Introduction}

Speakers do not always say exactly what they mean. For example, we might say a friend's poem ``wasn't terrible'' rather than saying ``it was bad'' to avoid hurting their feelings \cite{yoon20}, or just compliment specific elements that we liked without mentioning other elements we didn't like \cite{goffman2017interaction,pinker2008logic}. These kinds of politeness strategies allow speakers to balance competing goals, conveying accurate information while maintaining positive relationships \cite{hill1986universals,leech2014pragmatics}. As large language models (LLMs) are increasingly deployed in open-ended interactions across sensitive social domains like healthcare and education, their ability to appropriately use and understand polite language remains an important alignment challenge.

Politeness theory provides a valuable framework for addressing these questions. 
Seminal work by \citet{brown1987politeness} distinguishes between positive politeness strategies that affirm the listener (compliments, expressions of interest) and negative politeness strategies that minimize imposition (hedging, indirectness). 
While subsequent work has expanded this framework to encompass broader relational and rapport management concerns \citep{spencer2011conceptualising, watts2003politeness, locher2013relational}, this basic distinction remains crucial: different contexts call for different strategies, and mismatches can lead to communication breakdowns. 
If an AI system employs negative hedging strategies (``I am somewhat concerned that this approach might not be optimal'') in contexts where human speakers would expect positive, rapport-building strategies (``I love your creativity here, and wonder if we could build on it by...''), users may be left unsure whether the hedging reflects a genuinely negative evaluation or simply a systematic bias in expressions of politeness.

This kind of pragmatic misalignment represents a critical gap in our understanding of LLMs as social agents. While considerable attention has been paid to whether models can recognize politeness or generate polite language in constrained settings, effective social interaction depends not just on understanding politeness norms in the abstract but on actively selecting and applying appropriate strategies from a diverse linguistic repertoire. Human speakers navigate this complexity intuitively, deploying hedging, elaboration, indirect speech acts, and numerous other strategies to balance competing communicative goals in context-sensitive ways. To fully understand LLMs' grasp of politeness strategies, we need to examine whether they exhibit similar patterns of strategy selection and deployment across different contexts. This requires moving beyond limited-choice evaluations to examine open-ended language generation, where models have access to the full range of linguistic choices.
Rather than prescribing how AI systems ought to handle politeness, we seek descriptive evidence about current patterns of behavior.

In this work, we investigate we compare human and LLM politeness strategies in English-language scenarios, making the following contributions:
\begin{itemize}
\item We test whether LLMs reproduce human patterns of goal sensitivity in polite feedback using constrained response sets from \citet{yoon20}.
\item We collect and analyze a new dataset of open-ended responses from both humans and LLMs to identical social scenarios, enabling direct comparison of politeness strategies.
\item We perform detailed linguistic analyses to identify systematic differences in how humans and LLMs deploy various categories of politeness strategies.
\end{itemize}

Our results reveal that while LLMs have acquired important aspects of human-like pragmatic competence in polite language production – enough to be preferred by human evaluators – they also show systematic differences in strategy deployment that raise intriguing questions about the mechanisms underlying their social language capabilities. In particular, we find that models disproportionately rely on negative politeness strategies (minimizing imposition) even in contexts where humans prefer positive politeness strategies (building rapport), suggesting important differences in how these systems navigate social interactions\footnote{Data \& Code: \url{https://github.com/haoranzhao419/politeness-speech-production}}.

\begin{figure*}[ht]
\centering
\includegraphics[width=1.0\textwidth]{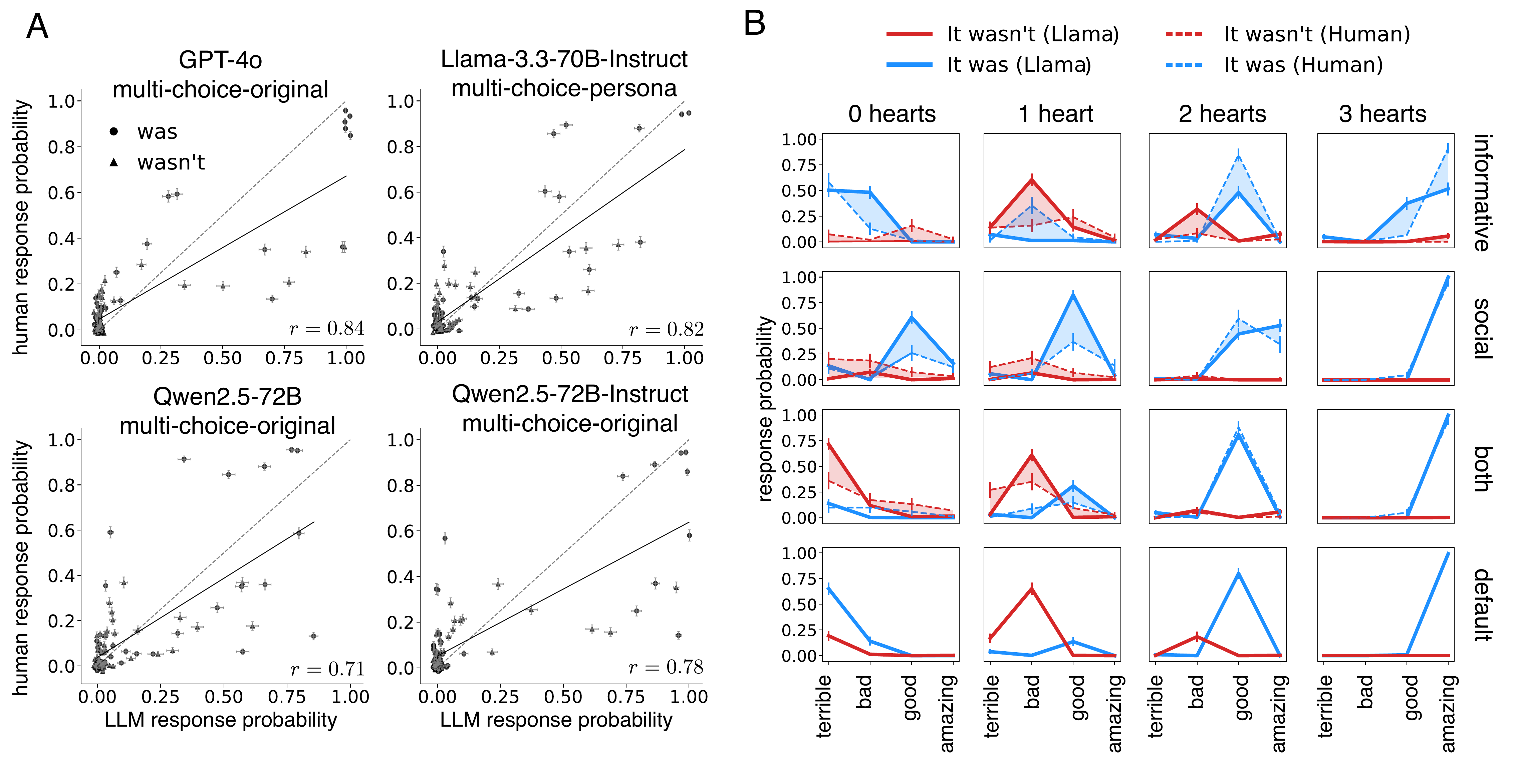}
\caption{(A) Correlations between human and model response probabilities for the top 4 models with specific prompting strategies we tested. Both the base and instruct-tuned versions of \texttt{Qwen2.5-72B} are shown here for comparison. Error bars are 95\% confidence intervals across vignettes. (B) Comparing the pattern of human and LLM responses across different communicative goals and ratings. Model results are from \texttt{Llama-3.3-70B-Instruct} using the multi-choice-persona prompting strategy; human responses are from \protect\citet{yoon20}.}
\label{fig:llm experiment results}
\end{figure*}

\section{Related Work}


\subsection{Computational models of politeness}

Research on politeness in linguistics and cognitive science has evolved from descriptive frameworks to quantitative models of pragmatic language use. 
Foundational work by \citet{brown1987politeness} established a systematic taxonomy of strategies, which has provided conceptual scaffolding for subsequent computational approaches. 
Studies in computational linguistics have since documented various linguistic markers of politeness across languages and contexts, examining formal features such as hedging, indirectness, and specific syntactic constructions correlate with perceived politeness \cite{danescu-niculescu-mizil-etal-2013-computational,aubakirova2016interpreting}.

Recent models in the Rational Speech Act (RSA) framework have explained the use of polite language as emerging from tradeoffs between informational utility capturing the desire for accuracy, a social utility representing the goal of making listeners feel good, and a self-presentational term reflecting speakers' desire to be perceived as both kind and honest \cite{yoon20,lumer2022modeling, carcassi2023handle, gotzner2024role}. 
This body of work has established a solid theoretical foundation for analyzing politeness as a pragmatic phenomenon arising from underlying tradeoffs. 
However, existing models have primarily focused on explaining choices among a small number of constrained utterance alternatives rather than modeling the rich variety of strategies humans employ in open-ended generation contexts.

\subsection{Pragmatic capabilities in LLMs}
Recent research has explored various aspects of pragmatic competence in large language models. Studies have examined LLMs' ability to understand indirect speech acts \cite{ruis2024goldilocks,jian2024llmsgoodpragmaticspeakers}, recognize conversational implicatures \cite{hu2022fine,lipkin2023evaluating}, and interpret non-literal language \cite{yerukola-etal-2024-pope,liularge}. These investigations predominantly employ multiple-choice formats, presenting models with pragmatic puzzles and evaluating their ability to select contextually appropriate interpretations. Results generally suggest that modern LLMs demonstrate sophisticated pragmatic understanding, often approaching human-like performance on benchmark tasks. However, these studies primarily assess recognition rather than production capabilities, leaving open questions about whether models can actively deploy pragmatic strategies in their own generated outputs.

\subsection{Polite language generation in LLMs}
Work on generating polite language in AI systems represents a small but growing research area. Early approaches focused on style transfer, with systems like those developed by \citet{niu-bansal-2018-polite} demonstrating that neural models could transform neutral text into more polite versions through specific syntactic transformations. Subsequent work explored paraphrasing to increase politeness \cite{fu2020facilitating}, politeness-focused style transfer \cite{madaan2020politeness}, and creating polite chatbots \cite{mukherjee2023polite}. However, these systems typically focused on surface-level transformations rather than strategic deployment of politeness based on contextual factors.
As noted in a recent survey \cite{priya2024computational}, existing approaches to polite language generation have predominantly emphasized isolated features (hedging expressions, please markers, specific lexical choices) rather than examining the full repertoire of politeness strategies and how they're selected based on communicative context. This leaves a significant gap in our understanding of whether LLMs can approximate the context-sensitivity that characterizes human politeness. Our work addresses this gap by directly comparing politeness strategies in humans and LLMs across varying communicative goals, examining whether models align with human preferences for positive versus negative politeness strategies in different contexts.

\section{Experiment 1: Constrained Settings}
\label{sec: constrained setting}

\begin{table*}[ht]
\centering
\label{tab:humanVllm}
\renewcommand{\arraystretch}{1.2}
\begin{tabular}{l|cc|cc|ccc}
& \multicolumn{4}{c|}{\textbf{Comparison with humans}} & \multicolumn{3}{c}{\textbf{Comparison with default goal}} \\
\toprule
\multirow{2}{*}{\textbf{LLMs}} & \multicolumn{2}{c|}{\textbf{Spearman}} & \multicolumn{2}{c|}{\textbf{MSE}} & \multirow{2}{*}{\textbf{vs. Both}} & \multirow{2}{*}{\textbf{vs. Inf.}} & \multirow{2}{*}{\textbf{vs. Social}} \\ 
& Original & Persona & Original & Persona & & & \\
\midrule
GPT-4o & \textbf{0.75} & \textbf{0.76} & 0.026 & 0.031 & 0.62 & \textbf{0.99} & 0.31 \\
Claude-3.5-Sonnet & 0.41 & 0.47 & 0.048 & 0.046 & \textbf{0.73} & 0.49 & 0.19 \\
\midrule
Llama-3.1-8B & 0.11 & 0.15 & 0.052 & 0.052 & 0.77 & \textbf{0.86} & 0.71 \\
Llama-3.1-8B-Instruct & 0.17 & 0.17 & 0.061 & 0.063 & \textbf{0.87} & 0.75 & 0.78 \\
Llama-3.1-70B & 0.66 & 0.67 & 0.034 & 0.030 & \textbf{0.86} & 0.58 & 0.57 \\
Llama-3.1-70B-Instruct & 0.73 & 0.74 & 0.023 & 0.024 & 0.74 & \textbf{0.75} & 0.53 \\
Llama-3.3-70B-Instruct & 0.67 & 0.66 & \textbf{0.018} & \textbf{0.019} & \textbf{0.80} & 0.64 & 0.40 \\
Mixtral-8x7B & 0.36 & 0.35 & 0.043 & 0.044 & 0.74 & \textbf{0.83} & 0.19 \\
Mixtral-8x7B-Instruct & 0.43 & 0.39 & 0.080 & 0.082 & \textbf{0.54} & 0.41 & 0.10 \\
Qwen2.5-72B & 0.65 & 0.66 & 0.028 & 0.029 & \textbf{0.83} & 0.75 & 0.73 \\
Qwen2.5-72B-Instruct & 0.66 & 0.64 & 0.033 & 0.034 & \textbf{0.63} & 0.55 & 0.54 \\
\bottomrule
\end{tabular}
\caption{Spearman correlations between the frequencies of human responses \protect\citep{yoon20} and LLM responses across all goal-rating combinations. Bold values indicate the highest correlation.}
\label{tab:goal-comparison}
\end{table*}

To what extent are LLMs sensitive to the goals that give rise to politeness in human speech? To address this question, we first examined whether LLMs could reproduce the patterns of goal-sensitive language use reported by \citet{yoon20}. Their study provided empirical evidence for a computational model of politeness where speakers strategically balance informational accuracy with social goals. Most notably, they found that when giving negative feedback, humans often deploy indirectness through negation \citep[e.g., ``wasn't terrible'' rather than ``bad''; see also][]{gotzner2024role,lumer2022modeling}.

We reimplemented this experiment with LLMs to assess their pragmatic competence in a constrained setting. In each scenario, a character gives feedback about another character's performance (e.g., a piano play or presentation), with the true quality ranging from 0 to 3 hearts. 
The speaker has one of four communicative goals: to be \emph{informative}, to be \emph{kind}, or \emph{both}. 
We also added a \emph{default} condition with no explicit goal specified to understand how LLMs behave by default. 
Models selected from the same set of eight responses used by humans, combining either ``was'' or ``wasn't'' with four adjectives (terrible, bad, good, amazing).

We tested a range of open-source (8B-72B parameters) and closed-source models using two prompting strategies: an ``original'' strategy that presented scenarios verbatim, and a ``persona'' variant that systematically varied speaker characteristics (e.g., gender, occupation, background) to better approximate the diversity in the population of human participants \cite{murthy-etal-2025-one,he2024psychometric}. For each model and prompting strategy, we sampled 30 responses with temperature $\tau = 1.0$ per scenario (see Appendix \ref{sec:Exp1} for details of prompt design).

\subsection{Model comparison}
We report Spearman correlation and mean squared error (MSE) between LLM and human responses as an overall measure of fit (see \autoref{tab:goal-comparison}). These results suggest that model size plays a crucial role in capturing human-like politeness strategies. Smaller models (\texttt{Llama-3.1-8B}) showed essentially no correlation with human responses, often failing to perform the multi-choice task at all, while intermediate-sized models like \texttt{Mixtral-8x7B} (effective model-size is $13$B \cite{jiang2024mixtralexperts}) showed only modest correlations.
However, larger models ($\ge$70B parameters) demonstrated much stronger alignment with human behavior, with \texttt{Llama-3.3-70B-Instruct} achieving the highest correlations among open-source models (Spearman $r = 0.67$). Among closed-source models, \texttt{GPT-4o} displayed particularly strong performance (Spearman $r = 0.75$), while \texttt{Claude-3.5-Sonnet} lagged behind with more modest correlations ($r = 0.41$). These findings suggest that sophisticated pragmatic competence for politeness emerges primarily in larger models, potentially reflecting the greater contextual sensitivity needed to balance competing communicative goals.

\subsection{Error analysis}
Despite strong overall correlations (see \autoref{fig:llm experiment results}A), even the best-performing models showed systematic differences from human responses. To better understand these patterns, we conducted a detailed comparison with human responses following the visualization approach in \citet{yoon20}.
The results in \autoref{fig:llm experiment results}B show that the best-fitting open-source model captures many key features of the human response patterns. Most notably, when rating a poor performance (0/3 hearts) with both informational and social goals, the model appropriately deploys negation as a politeness strategy, just as humans do: both humans and LLMs prefer to say ``wasn't terrible'' rather than ``was bad''. The model also closely tracks human preferences for positive ratings (2-3 hearts), showing appropriate sensitivity to the quality of the performance.

However, key differences emerged in the granularity of responses. Where humans show graded preferences across response options (distributing probability mass across multiple choices), LLMs tend toward more categorical binary choices, either strongly preferring or completely avoiding certain responses. They consistently choose one single option given a context, rating, and goal combination in most cases—despite our efforts to increase response diversity through temperature sampling ($\tau = 1.0$) or persona variation in prompting.

Closer analysis also revealed systematic differences in how well LLMs captured human behavior across different communicative goals. Models showed stronger alignment with human responses for the \emph{social} goal but underperformed when the goal was to be purely \emph{informative}. For example, when humans prioritize being informative about a poor performance, they often select direct negative feedback (``was bad''), while LLMs sometimes persist with softened language. This pattern suggests that, while LLMs have acquired some aspects of sophisticated politeness strategies, they may overapply these strategies even when directness would be more appropriate, potentially reflecting their training to be generally ``helpful and harmless''.

\subsection{Default goal analysis}

We included a \emph{default} goal condition (no explicitly specified goal) to evaluate how LLMs respond without specific communicative instructions. This condition helps reveal the implicit goals that might have been induced through various stages of model training. Although the overall fit to human data varies across models, we can ask which explicit goal produces the closest response pattern to the default goal, as measured by Spearman correlation.

Overall, we find a stronger resemblance to the \emph{both} goal (see \autoref{fig:llm experiment results}B), suggesting that models generally attempt to balance informativeness and social considerations by default. However, \autoref{tab:goal-comparison} reveals varying correlation patterns across different LLMs. While most models show stronger correlations with the \emph{both} goal, others correlate more strongly with the \emph{informative} goal. For instance, \texttt{Llama-3.3-70B-Instruct} appears to implicitly align with \emph{both} (Spearman $r = 0.80$), whereas \texttt{GPT-4o} shows much stronger alignment with \emph{informative} (Spearman $r = 0.99$).

These varied patterns suggest that the implicit goals guiding different LLMs' polite speech may reflect differences in their training objectives and alignment procedures. The dominant pattern of alignment with the \emph{both} goal is consistent with the general instruction to models to be both helpful (informative) and harmless (socially appropriate). However, the variability across models indicates that while these systems have acquired sophisticated politeness capabilities, the specific ways they balance competing goals may differ from model to model.

\section{Experiment 2: Open-ended Generation}

While our multiple-choice experiment demonstrated that larger LLMs can reproduce basic patterns of goal-sensitive politeness strategies, such as the strategic use of negation, this constrained format limits our understanding of how models deploy politeness in naturalistic settings. In real-world interactions, speakers draw from a rich repertoire of linguistic devices beyond those provided in fixed-choice scenarios. This raises a critical question: how do LLMs perform when given the freedom to generate polite language from scratch?

To address this question, we designed an open-ended generation experiment that uses the same scenarios as our multiple-choice study but removes the response constraints. This approach allows us to examine whether LLMs employ a similarly diverse and context-sensitive set of politeness strategies as humans when both have access to the full expressivity of language, and directly compare to results in the constrained setting.

\begin{figure*}[h!]
    \centering
    \includegraphics[width=1\textwidth]{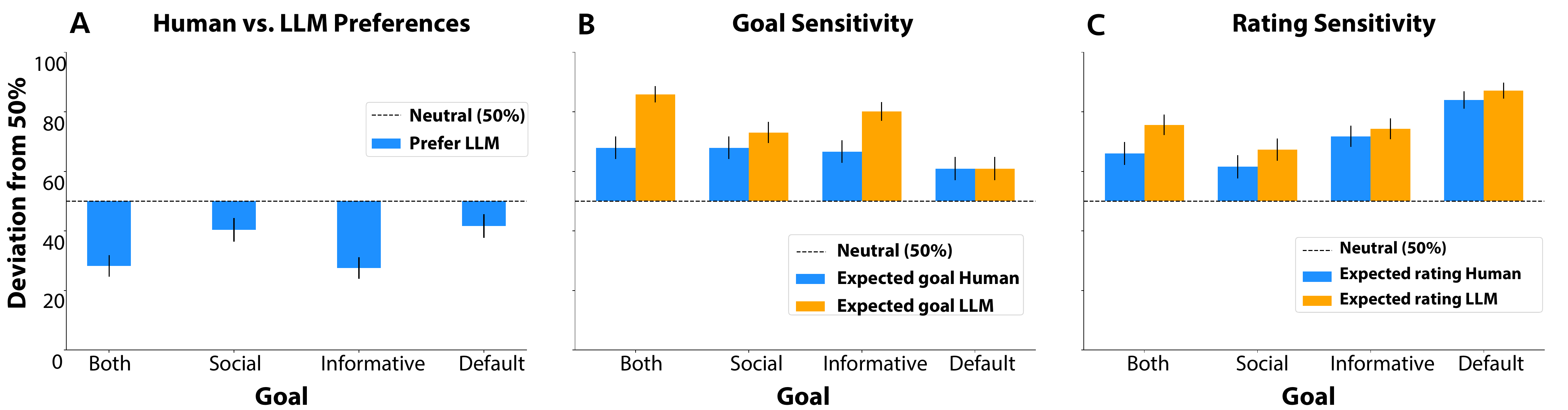} 
    \caption{Human evaluation results. The bars show the relative preference (50\% is chance). Bars above the 50\% line indicate the percentage to which responses are preferred as expected, and below indicate the percentage to which responses are preferred as unexpected. (A) Evaluators systematically prefer LLM generations over human generations. (B) Both humans and LLMs are sensitive to goals and (C) ratings. Error bars are bootstrapped 95\% confidence intervals.}
    \label{fig:human-eval-bar-plot}
\end{figure*}

\subsection{Methods}
\label{subsec:open-ended-design}
We used the scenarios from \citet{yoon20}, preserving the same performance ratings (0-3 hearts) and communicative goals (informative, social, both, default). We collected 3 open-ended responses per scenario from 156 human participants via Prolific (each responding to 4 distinct scenarios) and three responses from LLMs that performed well in our first experiment: \texttt{GPT-4o}, \texttt{Claude-3.5-Sonnet}, and \texttt{Llama-3.3-70B-Instruct}. We selected these larger models after calibration experiments revealed that smaller models failed to generate coherent responses in the open-ended format (see Appendix \ref{sec: calibartion}). Models were instructed to ``keep responses short and concise'' to ensure comparable length with human responses.

To assess preferences for these responses, we then conducted a two-alternative forced-choice evaluation with 156 human evaluators, each viewing four different scenarios. Evaluators made five judgments per scenario: (1) comparing human vs. LLM responses, (2-3) comparing goal-congruent vs. goal-incongruent responses for both sources, and (4-5) comparing rating-congruent vs. rating-incongruent responses for both sources. We randomized presentation order and ensured evaluators saw responses from different sources across blocks (see Appendix \ref{sec:Exp2} for full details).

\subsection{Results}
\label{subsec: analysis in the wild}


\paragraph{Overall preferences}

Surprisingly, human evaluators showed a marked preference for LLM-generated responses over human-generated ones across all goal types (66\% of all trials; see \autoref{fig:human-eval-bar-plot}A).
A mixed-effects logistic regression containing random intercepts at the evaluator and item level confirmed this preference was significantly different from chance ($z = 7.63$, $p < 0.001$).
This pattern held for each of the four communicative goals, with the largest effect observed for the informative goal (22\% above baseline) and the smallest effect observed for the default goal (8.3\% above baseline; see Figure~\ref{fig:human-eval-bar-plot}A).
However, there were systematic differences in the strength of these preferences across goals; a model including a fixed effect of goal accounted for significantly more variance than the intercept-only model, according to a likelihood-ratio test $\chi^2(3)=12.54, p = 0.006$.

\paragraph{Goal sensitivity}
Next, we considered the extent to which human-generated and LLM-generated utterances were goal-sensitive by calculating the proportion of trials where participants preferred a congruent utterance (i.e., an utterance actually produced to achieve the given goal) over an incongruent utterance (i.e., one produced under a different goal).
We found that both humans and LLMs demonstrated sensitivity to communicative goals: evaluators preferred the goal-congruent human response 15.9\% above-baseline ($z=6.89, p~<~0.001$), and preferred the goal-congruent LLM response even more strongly at 25.0\% above baseline $(z=9.59, p<0.001$).
Moreover, \autoref{fig:human-eval-bar-plot}B suggests that LLMs maintained greater or equal goal sensitivity across all four goals, indicating they successfully tailored their language to the specified communicative objective.

\paragraph{Rating sensitivity}
Finally, as a sanity-check, we asked whether utterances were sensitive to the actual state of the speaker (i.e., the number of hearts they felt about the performance being evaluated).
Again, both groups showed strong rating sensitivity. Human responses achieved a 20.8\% above-baseline preference for aligned ratings ($z = 8.32, p <0.001$), while LLM responses demonstrated even higher sensitivity with a 26.1\% above-baseline preference ($z = 9.59, p < 0.001$).
As shown in \autoref{fig:human-eval-bar-plot}C, LLMs maintained equal or improved sensitivity across all goals, indicating that they are not simply producing generically polite utterances but are modulating their responses appropriately as a function of both the basic information to be conveyed (the rating) and the specified communicative goal (e.g., being informative vs. making someone feel good).

\section{Linguistic Analysis of Politeness}

\begin{figure*}[h!]
    \centering
    \includegraphics[width=1\textwidth]{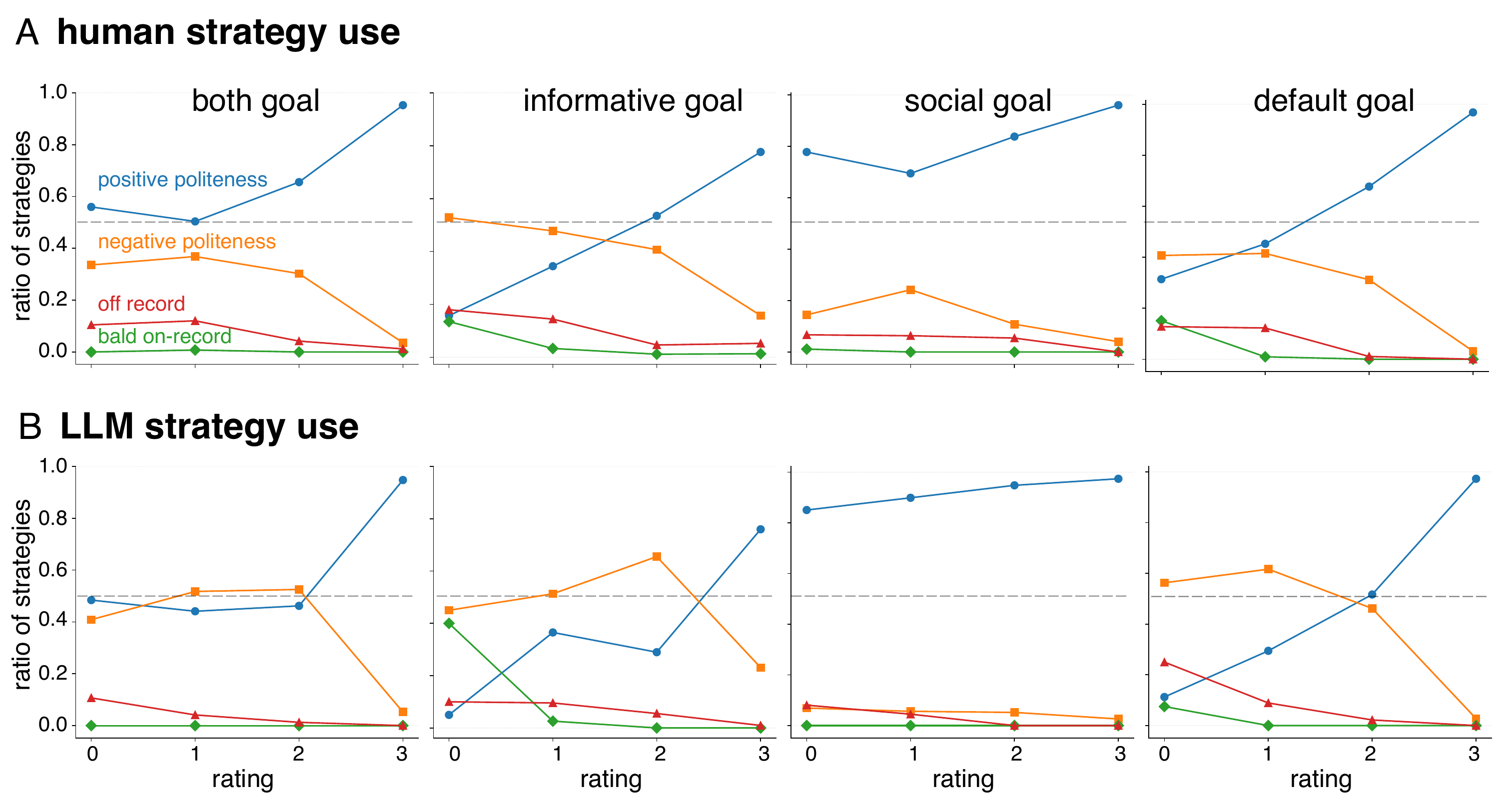} 
    \caption{Proportion of different politeness strategies  across ratings and goals for (A) human and (B) LLMs.}
    \label{fig:strategies-ratio-goal-rating}
\end{figure*}

While our evaluations show that LLMs successfully generate polite language that human evaluators prefer, these preferences alone don't reveal whether models use the same linguistic mechanisms as humans. To understand the specific politeness strategies employed by both humans and LLMs, we conducted a detailed linguistic analysis of the open-ended responses.

\subsection{Negation}

As a first step in our analysis, we examined how frequently the strategic use of negation documented in \citet{yoon20} and tested in Experiment 1 is employed in open-ended responses.
Among all 1,248 responses collected, 527 (42.2\%) used the specific pattern of adjective evaluation studied by \citet{yoon20}.
Within this subset, 35 responses (6.6\%) employed negation as a politeness strategy, and negation was most common in low-rating (0 or 1 heart) scenarios, which qualitatively replicates our findings from Experiment 1 (see Figure~\ref{fig:yoon-tessler analysis} in Appendix for details). 
Thus, the negation strategies studied in constrained settings do appear in open-ended production, but represent just one of many politeness devices available to speakers. 


\begin{table*}[t]
\centering
\begin{tabular}{lll}
\toprule
\textbf{Category} & \textbf{Politeness Strategy} & \textbf{Examples} \\
\midrule
\multirow{2}{*}{Positive Politeness} 
 & \textbf{Gratitude} & I'm so \textit{grateful} that \ldots \\
 & \textbf{Be optimistic} & You can even make them better next time! \\
\midrule
\multirow{2}{*}{Negative
Politeness} 
 & \textbf{Apologizing} & I didn’t like it, sorry! \\
 &\textbf{Question or hedge} & Maybe you can try something different? \\
 \midrule
\multirow{2}{*}{Off-Record} 
 & \textbf{Be vague} & It was \textit{interesting}. (true rating is 0) \\
 &\textbf{Give association clues} & It was better than those who can’t play.\\
 \midrule
\multirow{2}{*}{Bald on-Record} 
 & \textbf{Negative lexicon} & It was \textit{terrible}! \\
 & \textbf{Factuality} & I didn’t like the cookies at all. \\
\bottomrule
\end{tabular}%
\caption{Examples of utterances for different politeness strategies (two per category), with the corresponding sub-strategy highlighted in bold. See the comprehensive list in \autoref{tab:politeness-strategies-list}.}
\label{tab:politeness-strategies-list-example}
\end{table*}

\subsection{Word usage patterns}

To better understand differences between human and LLM responses, we analyzed unigram distributions using Pointwise Mutual Information (PMI) and Jensen-Shannon Divergence (JSD).
First, examining unigrams with the highest PMI, we found that human responses more frequently incorporated casual language and expressions (e.g., ``awesome," ``great"), whereas LLM-generated responses tended toward more formal linguistic choices (e.g., ``fabulous," ``excellent"). Both groups effectively employed personalization as a politeness strategy, such as directly mentioning the performer's name (e.g., ``Your app is pretty good, Henry!"). Additionally, both humans and LLMs adapted their lexical choices based on context, with minimal overlap in high-PMI words across different goals and ratings.

Next, we quantified differences in empirical word frequency distributions by calculating the Jensen-Shannon Divergence (JSD). 
Interestingly, the JSD between the lexical distributions of preferred and non-preferred response groups was quite small (JSD = 0.013) though still significantly different than a permuted null distribution ($p < 0.001$), while all other group comparisons showed much larger differences (JSD > 0.13, $p < 0.001$; see \autoref{tab:jsd-word-frequency}). This suggests that simple lexical choice may not be the primary driver of human preferences in polite language.

Finally, we conducted higher-dimensional analyses using SBERT embeddings \cite{sbert-2019} to distinguish between response categories. These analyses (described in Appendix \ref{sec:text-classifiers}) revealed that while human vs. LLM responses were readily distinguishable in embedding space (83\% accuracy), preferred vs. non-preferred responses were much harder to classify (54\% accuracy). 
LLM responses were more distinguishable across different communicative goals than human responses, suggesting more stereotyped strategies.



\subsection{Annotated politeness strategies}

To obtain a comprehensive picture of the politeness strategies employed in human and LLM responses, we conducted a detailed annotation using the politeness framework from \citet{brown1987politeness}, supplemented by markers from \citet{danescu-niculescu-mizil-etal-2013-computational}. 
This framework distinguishes four broad categories of politeness strategies with many subtypes (see \autoref{tab:politeness-strategies-list-example} and Appendix \autoref{tab:politeness-strategies-list}): positive politeness (e.g. compliments and expressions of interest), negative politeness (e.g. hedging and indirectness), off-record strategies (indirect hints that maintain plausible deniability), and bald-on-record strategies (direct statements without politeness).
We used LLMs (\texttt{GPT-4.1} and \texttt{Claude-3.7-Sonnet}) as annotators, following best practices \cite{tan2024large}. Annotations were manually verified and corrected where necessary. We used LLMs (\texttt{GPT-4.1} and \texttt{Claude-3.7-Sonnet}) as annotators, following best practices \cite{tan2024large}. Annotations were manually verified and corrected where necessary. Appendix \ref{sec:politeness-annotation} provides full details on the annotation process and framework.

\paragraph{Overall strategy distribution.} 
As shown in Figure~\ref{fig:strategies-ratio-goal-rating}, both humans and LLMs rely primarily on positive politeness (rapport-building) and negative politeness (minimizing imposition) strategies, with relatively low use of off-record and bald-on-record approaches. However, a key difference emerged in strategy selection patterns: while both humans and LLMs increased their use of positive politeness strategies as ratings increased, LLMs showed systematically higher use of negative politeness strategies even in positive contexts (higher ratings), where humans tended to reduce such strategies.
These strategy distributions were significantly different under a permutation test (JSD = 0.023, $p < 0.001$). 
This pattern, where LLMs maintain high levels of hedging and indirectness across contexts, may reflect training objectives that prioritize avoiding potential harm over more human-like affirmation strategies.

\paragraph{Goal and rating sensitivity.} 
We observed that both humans and LLMs appropriately varied their strategy distribution by communicative goal, with the informative goal showing the most distinct pattern. For the informative goal with high ratings (2-3 hearts), LLMs showed unexpectedly higher use of negative politeness strategies compared to humans, who shifted toward positive strategies in these contexts. This pattern suggests that LLMs may overuse hedging, conventional indirectness, and other distancing strategies even when giving positive feedback, potentially explaining some of the stylistic differences observed in the evaluation.
However, as with word usage patterns, the differences between strategy distributions for preferred and non-preferred responses were not significant (JSD = 0.008, $p = 0.087$), suggesting that preference judgments may be driven by higher-order social factors beyond the mere presence or absence of specific words or politeness strategies.

\paragraph{Cross-cultural variation.}
Given documented cultural differences in politeness norms, we also conducted an exploratory analysis testing whether strategy distributions varied between our US and UK participants. We found a statistically significant difference (JSD = 0.5279, $p<0.001$). US participants deployed a more diverse range of sub-strategies overall than UK participants, especially among positive strategies; however, the effect size was relatively small and our sample was imbalanced (532 US responses vs. 92 UK responses). While this preliminary finding suggests some cross-cultural variation even within English-speaking populations, more balanced sampling would be needed to characterize these differences systematically. Future work should examine how cultural context shapes both human and LLM politeness strategies across a broader range of English varieties and other languages.

\section{Discussion}

We find that LLMs demonstrate impressive pragmatic competence in capturing politeness phenomena, but also differ in important ways from humans. 
Most notably, in open-ended production, LLMs continue to rely on negative politeness strategies (hedging, indirectness) even in positive contexts where humans shift toward positive strategies (rapport-building). 
Interestingly, LLM responses were consistently preferred by human evaluators over crowd-sourced human responses, suggesting that preference judgments may track different features of the utterance than the mere ability to reproduce human-like patterns.
This contrast --- aligning with human patterns in constrained multiple-choice tasks while diverging in more open-ended tasks --- emphasizes that LLMs leverage a richer repertoire of strategies when given the opportunity to express them, such as the familiar ``\textsc{compliment}, but \textsc{criticism}'' constructions \cite{prochazka2020sandwich} we observed in low-rating scenarios, which were not provided as available options in the constrained tasks.

Recent developments in politeness theory may offer insight into these results. 
While \citet{brown1987politeness} classically emphasized the role of face threat, subsequent work has reconceptualized politeness as part of a broader class of \emph{relational work} \citep{spencer2011conceptualising,locher2013relational}.
From this perspective, politeness strategies manage the affective quality of relationships through multiple components: face concerns (desires for positive evaluation and acknowledgment of identity), sociality rights (what people expect from each other in interaction, including expectations about appropriate levels of imposition and warmth), and other relational goals.

Our findings suggest that humans and LLMs may functionally assign different weights to these components. 
Humans emphasize equity rights (avoiding imposition) when delivering criticism but shift toward association rights (building warmth) when giving praise.
But LLMs consistently maintain a high weight on imposition across all contexts, potentially aligning with user expectations for AI systems, where maintaining appropriate distance could be more desirable than pursuing interpersonal warmth. 
Understanding how different training objectives and alignment procedures implicitly shift these relational weights represents an important direction for future work, particularly so developers can make intentional choices about more or less desirable interpersonal dynamics.

Distributional differences in politeness strategies may also have practical implications for human-AI communication. 
Following Gricean principles of communication as rational social action \cite{Grice1989-GRILAC}, listeners expect speakers to modulate their politeness strategies based on the context: you shouldn't need to take steps to mitigate face threat if face isn't threatened. 
So when AI systems violate these expectations, they create an interpretive puzzle: users must determine whether hedged language reflects the literal content being communicated or whether it is an artifact of the system's persona, risking pragmatic misinterpretations where humans interpret hedged positive feedback as more negative than intended. For example, does ``your analysis seems reasonably sound'' mean there are specific weaknesses that ought to be addressed, or is this just how the system expresses unqualified approval?
Users may adapt their expectations of the system over time \citep{branigan2010linguistic,araujo2024speaking,zhou2023talking,waytz2014mind}, but may still incur a hidden cognitive cost associated with correcting for these differences.
Future work should examine how these differences play out across cultural contexts, where politeness norms already vary substantially \citep{wierzbicka2003cross,ide1989formal}.

In conclusion, our findings provide an empirical foundation for understanding differences between humans and AI systems in a key domain of pragmatic language use.
Critically, our study was descriptive rather than prescriptive.
Divergences matter not because artificial simulacra of human-like behavior is desirable, but because it affects human-AI communication in practice, raising questions about how users learn to interpret AI-generated politeness differently than human-generated politeness.
Beyond practical applications, these findings also contribute to theoretical debates about the cognitive architecture underlying polite speech, and the computational principles that guide decisions across social contexts.
These findings could inform future LLM training approaches to better align with human pragmatic patterns, laying groundwork for making intentional rather than accidental choices about subtle pragmatic patterns in AI systems.

\section*{Limitations}
While our work gave a comprehensive picture of comparing the polite language use in humans and LLMs, there are still limitations that could be addressed in future work. 
First and foremost, we want to point out that our study exclusively focused on English-language politeness strategies. While we acknowledge that politeness strategies vary largely across languages and cultures, cross-linguistic and cross-cultural analyses are beyond the scope of this work and represent important directions for future research. 

Regarding the ``persona'' prompting in Experiment 1, we varied demographic factors (e.g., gender, occupation, etc.) across personas to increase response diversity. While demographic influences on politeness are not the focus of this study, and we observed no clear demographic effects on LLM performance, we acknowledge the potential risks that explicitly using demographic categories in prompts could activate stereotypes in LLM responses. One explanation from a recent study is that specific demographic information can activate LLM stereotypes through ``implicit personalization,'' where models automatically infer users' demographic attributes from subtle conversational cues (topics, language patterns, cultural references) and then generate responses based on those stereotypical associations \cite{neplenbroek2025readingpromptsstereotypesshape}. To mitigate this risk, future work could explore alternative methods for eliciting response variation, such as varying communicative contexts rather than speaker demographics. 
Additionally, in our human experiments, we acknowledge that the preferences given by human evaluators were from a third-person perspective, which may underestimate the effects of politeness strategies. For instance, receiving overly hedged criticism (``Your work is a bit lacking'') might feel more threatening or frustrating in a first-person context where it is your own work, while positive rapport-building strategies might be more appreciated by direct recipients. Future research should examine first-person reactions to these politeness patterns, ideally through synchronized interaction scenarios where participants receive feedback from other humans vs. LLMs, which would better capture the relational and emotional dimensions of pragmatic (mis)alignment.

Furthermore, throughout our analyses, we still cannot answer the question of what makes human evaluators prefer the responses they prefer, as all our analyses showed very minimal differences between preferred and non-preferred responses. One guess is that even ratings and goals are made very clear in the provided scenarios, human evaluators still may not pay enough attention to, and optionally omit this information, instead, they tend to pick whichever one in the given pair that sounds nicer. Future research, for example, testing LLMs as evaluators and comparing LLM-as-evaluator preference results with humans, could give us more insight into this question. Also, as our results show that LLMs are still not quite human-like in picking the right politeness strategies in a context-sensitive way, future research on how to develop computational methods and algorithms to make LLMs better at polite language use and as social agents will be necessary.


\section*{Acknowledgments}

We thank Takateru Yamakoshi for assistance with the LLM evaluation code, Yuka Machino for feedback on the draft, and Noah Goodman for early input and encouragement. We also would like to thank anonymous reviewers for providing valuable feedback.

\bibliography{custom}

\appendix

\section{Experiment 1 Methods}
\label{sec:Exp1}
We closely followed the experimental paradigm of \citet{yoon20}.
In this study, participants read short scenarios about someone seeking feedback on a performance or creative work. Each scenario specified (1) the true quality of the work on a scale from 0 to 3 hearts and (2) the speaker's communicative goal -- either to be informative, to make the person feel good, or to do both. 
Participants then chose what they would say from a restricted set of options, combining either \emph{was} or \emph{wasn't} with one of four adjectives: \emph{terrible}, \emph{bad,} \emph{good}, or \emph{amazing.} 
Scenarios were constructed from 13 different contexts (e.g., filmmaking, songwriting, concert performance), yielding 156 unique scenarios (13 contexts × 4 ratings × 3 goals). 
We also added a ``default" condition with no explicitly specified goal, bringing our total to 208 scenarios.

To test LLMs on this task, we developed two prompting strategies. In our basic approach, which we called ``multi-choice-original", we simply presented each scenario verbatim and asked the model to choose from the eight possible responses (all combinations of ``was"/``wasn't" with the four adjectives). To better approximate the diversity of human participants and with the hope to see that diversifying the personas of LLMs would improve their performance, we also considered a ``persona" variant where we systematically varied speaker characteristics like gender, occupation, and background, where we call ``multi-choice-persona". We tested these approaches across a range of current LLMs, including both closed-source (\texttt{GPT-4o}, \texttt{Claude-3.5-Sonnet}) and open-source models (\texttt{Llama-3}, \texttt{Mixtral}, \texttt{Qwen2.5}) of varying sizes (8B to 70B parameters). For open-source models, we compared both base and instruct-tuned versions where available to see the influence of the post-training stage on this task. To approximate the multiple participants in human studies, we collected 30 responses per scenario from each model using a temperature of $\tau = 1.0$.

\subsection{Prompting strategies}

The system prompt remained consistent between the original and persona prompting strategies.

\paragraph{Multi-choice-original/persona system prompt}: \texttt{You will see a scenario below.
In the scenario, person A is asking for person B's opinion on their performance.\texttt{\textbackslash n}
Person B's true feelings in the scenario are shown on a scale of 0 to 3 hearts.\texttt{\textbackslash n}
0 heart, the lowest rating, means the person does not like the performance at all, and 3 hearts, the highest rating, means the person likes it a lot.\texttt{\textbackslash n}
Please read the scenario carefully and answer the question ONLY with one of the eight options provided.\texttt{\textbackslash n}
Please provide your response in the following format:\texttt{\textbackslash n}
Answer:<one of the eight possible answer options in the scenario>}

To construct persona prompts, we varied the following details:
\begin{itemize}
    \item Race: \{white, Black, Asian, Hispanic, American-Indian\}
    \item Gender: \{woman, man, non-binary person\}
    \item City: \{New York, Chicago, San Francisco, Boston, Houston\}
    \item Years of experience: \{17, 18, 19, 20, 21, 22, 23\}
    \item Occupation: \{a critic, an expert, a teacher, a friend, a colleague, an acquaintance\}
\end{itemize}

\paragraph{NB:} We chose these demographic variables not because we hypothesized that these groups systematically differ in politeness strategies, but rather our use of personas was solely a technique to increase overall response diversity, attempting to approximate the natural variation we see in the population of human responses (as opposed to the modal ``original'' response). In our analysis, we explicitly did NOT analyze responses by demographic category, as our goal was simply to avoid the kind of mode collapse we observed in preliminary tests without personas. 

For example, here is an example of what the scenarios look like with and without a persona.

\begin{quote}
\textbf{Scenario without a persona:} Imagine that John just gave a presentation, but John didn't know how good it was. John approached Chris, who knows a lot about giving presentations, and asked ``How was my presentation?''

\textbf{Scenario with a persona:} Imagine that John just gave a presentation, but John didn't know how good it was. John approached Chris, \textit{an Asian man from Boston, who has 19 years of experience as a teacher in the field} and knows a lot about giving presentations. John asked ``How was my presentation?''
\end{quote}

A complete instance of our scenario as fed into the LLM user prompt looks like this — the example is a multi-choice-original, 2-hearts rating, informative goal scenario.
\begin{quote}

\textbf{Context:} Imagine that Bob just gave a presentation, but Bob didn't know how good it was. Bob approached John, who knows a lot about giving presentations, and asked ``How was my presentation?''

\textbf{Rating:}
Here's how John actually felt about Bob's presentation:
2 out of 3 hearts

\textbf{Question:}
If John wanted to give as accurate and informative feedback as possible, but not necessarily make Bob feel good, What would John be most likely to say?

\textbf{Options:}
\begin{enumerate}
    \item It was terrible.
    \item It was bad.
    \item It was good.
    \item It was amazing.
    \item It wasn't terrible.
    \item It wasn't bad.
    \item It wasn't good.
    \item It wasn't amazing.
\end{enumerate}
\end{quote}

\subsection{Additional results}

See \autoref{tab:human-llm-comparison} for a complete comparison between LLM and human response patterns across Spearman, Pearson, and MSE metrics using both multi-choice-original and multi-choice-persona prompting strategies, as a complement to \autoref{tab:goal-comparison} in the main text, where Pearson correlation scores were not reported.

See \autoref{tab:LLM-human-goals-comparison-pearson} and~\ref{tab:LLM-human-goals-comparison-spearman} for comprehensive comparison results between human and LLM responses across different goals. Since the ``default'' goal case was not studied in \citet{yoon20}, we focused on ``both'', ``social'', and ``informative'' goals and report both Pearson and Spearman correlation scores. For the two tables, we observed that different LLMs have medium to strong correlations with human responses. Both base and instruct-tuned versions of \texttt{Llama-3.1-8B} and \texttt{Mixtral-8x7B} showed very low correlation scores and their incompetence in generating polite language.

See \autoref{tab:goals-comparison-llm-pearson} and~\ref{tab:goals-comparison-llm-spearman} for a complete comparison report between ``default'' and other goals as a complement to \autoref{tab:goal-comparison} in the main text. We reported both Pearson and Spearman correlation scores for ``multi-choice-original'' and ``multi-choice-persona'' prompting strategies. We found that including personas does not have any real effect, and the findings are consistent with those reported in the main text.

\begin{table*}[h!]
\centering
\renewcommand{\arraystretch}{1.2} 
\begin{tabular}{lcccc|cc}
\toprule
\multirow{2}{*}{\textbf{LLMs}} & \multicolumn{2}{c}{\textbf{Pearson}} & \multicolumn{2}{c}{\textbf{Spearman}} & \multicolumn{2}{c}{\textbf{MSE}} \\ 
                  & Original & Persona & Original & Persona & Original & Persona \\ 
\midrule
GPT-4o   &  \textbf{0.84}     & 0.81     & \textbf{0.75}      & \textbf{0.76}   & 0.026 &  0.031 \\
Claude-3.5-Sonnet & 0.50     & 0.55     & 0.41      & 0.47   & 0.048 &  0.046 \\
\midrule
Llama-3.1-8B & -0.02    & -0.02    & 0.11   & 0.15   & 0.052    & 0.052    \\
Llama-3.1-8B-Instruct & -0.05   & -0.01     & 0.17     &  0.17  & 0.061    & 0.063    \\
Llama-3.1-70B & 0.59    & 0.63   & 0.66      & 0.67  & 0.034 & 0.030  \\
Llama-3.1-70B-Instruct & 0.76   & 0.74    &   0.73    & 0.74    &  0.023     & 0.024     \\
Llama-3.3-70B-Instruct & 0.83   & \textbf{0.82}     & 0.67  & 0.66     & \textbf{0.018}     &  \textbf{0.019}    \\
Mixtral-8x7B & 0.36    & 0.33     &  0.36    &  0.35   &  0.043     & 0.044   \\
Mixtral-8x7B-Instruct & 0.29    & 0.29     & 0.43     & 0.39     & 0.080     &  0.082    \\
Qwen2.5-72B  & 0.71    &  0.70   & 0.65    & 0.66    & 0.028      &  0.029    \\
Qwen2.5-72B-Instruct  & 0.78   &0.77      & 0.66    & 0.64   & 0.033     & 0.034    \\
\bottomrule
\end{tabular}
\caption{Complete version of model comparison results reported in Table 1 in the main text, including Pearson correlation scores.}
\label{tab:human-llm-comparison}
\end{table*}

\begin{table*}[h!]
\centering
\renewcommand{\arraystretch}{1.2} 
\begin{tabular}{lcccccc}
\toprule
\textbf{LLMs} & \multicolumn{2}{c}{\textbf{Both}} & \multicolumn{2}{c}{\textbf{Social}} & \multicolumn{2}{c}{\textbf{Inf.}} \\ 
              & Original & Persona & Original & Persona & Original & Persona \\ 
\midrule
GPT-4o                 & \textbf{0.89} & \textbf{0.89} & 0.80 & 0.75 & 0.83 & 0.78 \\
Claude-3.5-Sonnet      & \textbf{0.58} & \textbf{0.65} & 0.49 & 0.52 & 0.46 & 0.53 \\
\midrule
Llama-3.1-8B           & 0.05 & 0.01 & -0.09 & -0.03 & -0.03 & -0.04 \\
Llama-3.1-8B-Instruct  & 0.10 & 0.09 & -0.15 & -0.03 & -0.11 & -0.11 \\
Llama-3.1-70B          & 0.66 & 0.73 & \textbf{0.79} & \textbf{0.81} & 0.36 & 0.39 \\
Llama-3.1-70B-Instruct & \textbf{0.92} & \textbf{0.91} & 0.88 & 0.87 & 0.46 & 0.42 \\
Llama-3.3-70B-Instruct & \textbf{0.92} & \textbf{0.92} & 0.88 & 0.88 & 0.70 & 0.68 \\
Mixtral-8x7B           & 0.18 & 0.20 & 0.63 & 0.62 & 0.19 & 0.11 \\
Mixtral-8x7B-Instruct  & 0.32 & 0.19 & 0.62 & 0.84 & -0.08 & -0.10 \\
Qwen2.5-72B            & 0.83 & 0.73 & \textbf{0.85} & \textbf{0.83} & 0.47 & 0.57 \\
Qwen2.5-72B-Instruct   & \textbf{0.86} & \textbf{0.85} & 0.84 & 0.83 & 0.65 & 0.64 \\
\bottomrule
\end{tabular}
\caption{Pearson Correlation between human and LLMs over different goals}
\label{tab:LLM-human-goals-comparison-pearson}
\end{table*}

\begin{table*}[h!]
\centering
\renewcommand{\arraystretch}{1.2} 
\begin{tabular}{lcccccc}
\toprule
\textbf{LLMs} & \multicolumn{2}{c}{\textbf{Both}} & \multicolumn{2}{c}{\textbf{Social}} & \multicolumn{2}{c}{\textbf{Inf.}} \\ 
              & Original & Persona & Original & Persona & Original & Persona \\ 
\midrule
GPT-4o      &  0.70   &  0.71    &  0.74    & 0.74    & \textbf{0.80}    &  \textbf{0.80}  \\
Claude-3.5-Sonnet &  0.38     & \textbf{0.49}   & 0.43     & \textbf{0.47}  &  0.41  &0.41    \\
\midrule
Llama-3.1-8B & 0.12    & 0.10    & 0.16   & 0.15  & 0.02    &  0.15   \\
Llama-3.1-8B-Instruct & 0.32  & 0.39     &  0.17    &  0.17   & 0.06     &  0.06   \\
Llama-3.1-70B & 0.65    & 0.68   & 0.63      & 0.64   & \textbf{0.69}      &  \textbf{0.72}  \\
Llama-3.1-70B-Instruct & 0.66   &  0.74   & 0.67      & 0.67    &  \textbf{0.86}     & \textbf{0.83}    \\
Llama-3.3-70B-Instruct &0.64    & 0.68     &0.58   &  0.60    & \textbf{0.77}      & \textbf{0.72}     \\
Mixtral-8x7B & 0.33    &  0.35    &0.23      &0.21     &0.46       & 0.44    \\
Mixtral-8x7B-Instruct & 0.34    & 0.33     &0.33      & 0.17     & 0.60      &  0.67    \\
Qwen2.5-72B  & \textbf{0.70}   & 0.64    &0.59     & 0.58     & 0.69      & \textbf{0.74}     \\
Qwen2.5-72B-Instruct  & \textbf{0.69}   & 0.63     &0.63     & \textbf{0.66}    &0.64      &0.64     \\
\bottomrule
\end{tabular}
\caption{Spearman Correlation between human and LLMs over different goals}
\label{tab:LLM-human-goals-comparison-spearman}
\end{table*}

\clearpage

\begin{table*}[h!]
\centering
\renewcommand{\arraystretch}{1.2} 
\begin{tabular}{lcccccc}
\toprule
\textbf{LLMs} & \multicolumn{2}{c}{\textbf{vs. Both}} & \multicolumn{2}{c}{\textbf{vs. Inf.}} & \multicolumn{2}{c}{\textbf{vs. Social}} \\ 
              & Original & Persona & Original & Persona & Original & Persona \\ 
\midrule
GPT-4o      & 0.59    &  \textbf{0.66}    & \textbf{0.64}     &  \textbf{0.66}   &  0.43     & 0.40    \\
Claude-3.5-Sonnet &  \textbf{0.64}    &  \textbf{0.80}    & 0.50 & 0.41    & 0.05    &  0.04   \\
\midrule
Llama-3.1-8B & \textbf{0.76}    &  0.70   &  \textbf{0.83}  & 0.67  & 0.70    &  0.69   \\
Llama-3.1-8B-Instruct & \textbf{0.87}   &  \textbf{0.81}     & 0.57     &   0.66   & 0.72    &  0.74    \\
Llama-3.1-70B & \textbf{0.86}    &  \textbf{0.86} & 0.48      & 0.58    & 0.47      &  0.54  \\
Llama-3.1-70B-Instruct & \textbf{0.82}   & \textbf{0.89}    & 0.67      &  0.62   & 0.47      &  0.43   \\
Llama-3.3-70B-Instruct & 0.83   & \textbf{0.85}     & \textbf{0.92 }  & 0.82     & 0.49     &  0.52    \\
Mixtral-8x7B & 0.69     & 0.61     & \textbf{0.78}      &  \textbf{0.87}   &0.01       &  -0.07   \\
Mixtral-8x7B-Instruct & 0.28    &  0.30    & \textbf{0.45}      & \textbf{0.43}     &0.04      & 0.29     \\
Qwen2.5-72B  & \textbf{0.77}    &  \textbf{0.78}   & 0.66     &0.72     &0.70      & 0.73     \\
Qwen2.5-72B-Instruct  &0.84    & \textbf{0.86}      &\textbf{ 0.87}     &0.75    &0.57      &0.54     \\
\bottomrule
\end{tabular}
\caption{Pearson Correlation between default goal and other goals in Experiment 1}
\label{tab:goals-comparison-llm-pearson}
\end{table*}

\begin{table*}[h!]
\centering
\renewcommand{\arraystretch}{1.2} 
\begin{tabular}{lcccccc}
\toprule
\textbf{LLMs} & \multicolumn{2}{c}{\textbf{vs. Both}} & \multicolumn{2}{c}{\textbf{vs. Inf.}} & \multicolumn{2}{c}{\textbf{vs. Social}} \\ 
              & Original & Persona & Original & Persona & Original & Persona \\ 
\midrule
GPT-4o      & 0.62  & 0.46 &  \textbf{0.99} & \textbf{0.99}     &0.31      & 0.33   \\
Claude-3.5-Sonnet &  \textbf{0.73}     & \textbf{ 0.86}   & 0.49     & 0.63   & 0.19    & 0.44    \\
\midrule
Llama-3.1-8B &  0.77   & \textbf{0.72}    & \textbf{0.86}   &0.68   &0.71     &  0.67   \\
Llama-3.1-8B-Instruct & \textbf{0.87}  &  \textbf{0.82 }    &  0.75    &  0.71     & 0.78    & 0.72     \\
Llama-3.1-70B & \textbf{0.86}    &  \textbf{0.79}  & 0.58      & 0.73  & 0.57      &  0.58  \\
Llama-3.1-70B-Instruct & 0.74   &  \textbf{0.79}   & \textbf{0.75}      &  \textbf{0.79}   & 0.53      &  0.38   \\
Llama-3.3-70B-Instruct & \textbf{0.80}   & \textbf{0.76}     & 0.64  & 0.71     & 0.40     &  0.41    \\
Mixtral-8x7B &  0.74   & 0.64     & \textbf{0.83}      &  \textbf{0.84}   &0.19     & -0.03    \\
Mixtral-8x7B-Instruct & \textbf{0.54}    & \textbf{0.58}     &0.41      & 0.37     &0.10      &  0.08    \\
Qwen2.5-72B  & \textbf{ 0.83}    & \textbf{0.75}    & 0.75     &  0.74   &0.73      &  0.72    \\
Qwen2.5-72B-Instruct  & \textbf{0.63}    & 0.61     &0.55     &\textbf{ 0.64 }   &0.54     &0.52     \\
\bottomrule
\end{tabular}
\caption{Spearman Correlation between default goal and other goals in Experiment 1}
\label{tab:goals-comparison-llm-spearman}
\end{table*}

\clearpage

\begin{figure*}[h!]
    \centering
    \includegraphics[width=0.99\textwidth]{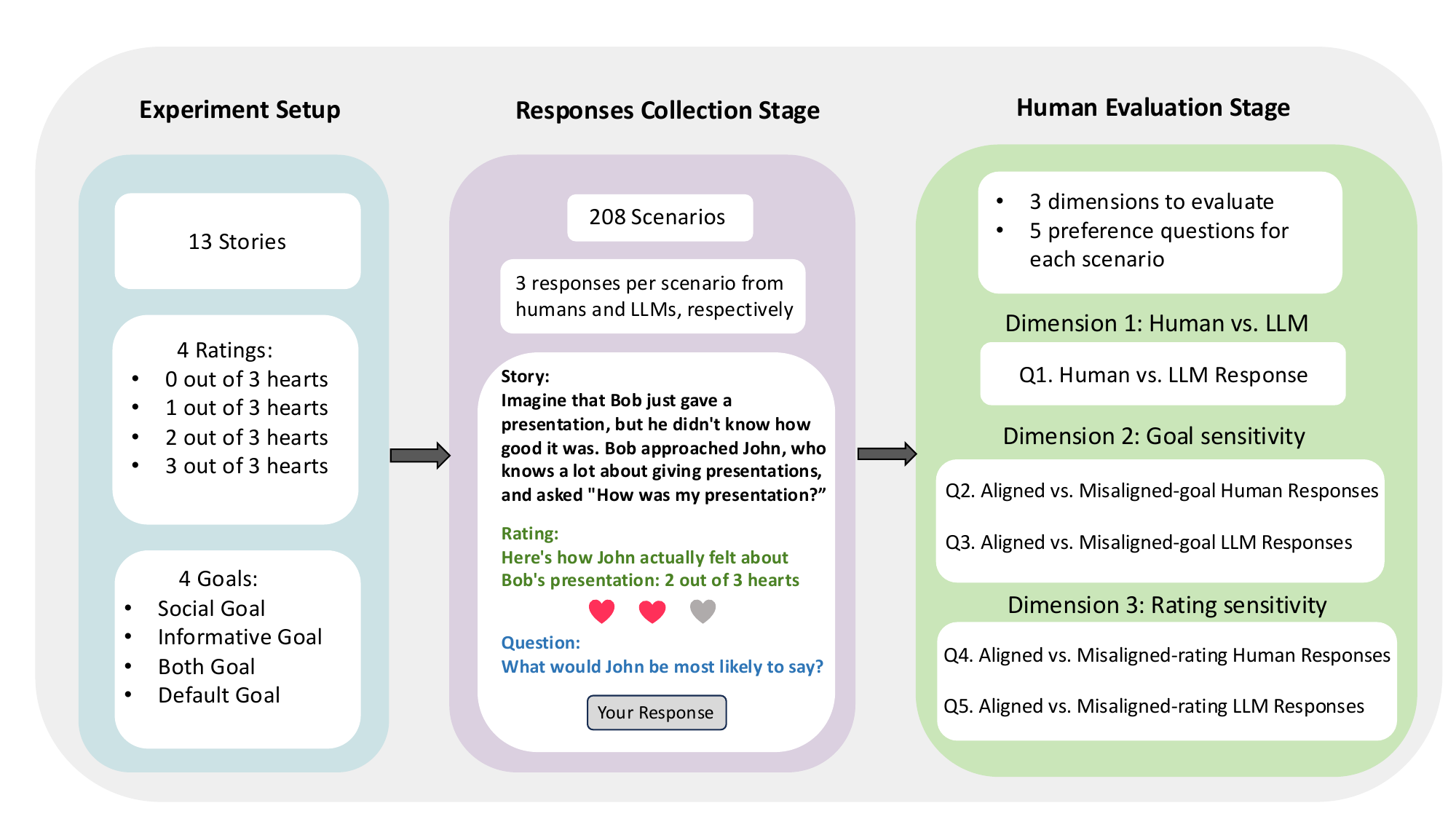} 
    \caption{Pipeline for comparing open-ended polite speech generation in humans and LLMs. Our study consists of two stages: an initial stage where we elicit responses for a variety of scenarios and a second stage where we ask a naive group to evaluation which of these responses they prefer.}
    \label{fig:experiment-pipeline}
\end{figure*}

\begin{table*}[h!]
\centering
\renewcommand{\arraystretch}{1.3}
\setlength{\tabcolsep}{8pt}
\begin{tabular}{>{\raggedright\arraybackslash}m{5cm} >{\raggedright\arraybackslash}m{5cm} >{\raggedright\arraybackslash}m{4.5cm}}
\toprule
\textbf{Scenario} & \textbf{Rating: 0/3 hearts, Goal: Both} & \textbf{Rating: 2/3 hearts, Goal: Informative} \\ 
\midrule
\multirow{2}{=}{\parbox{5cm}{Imagine that Jenny wrote a poem, but she didn't know how good it was. Jenny approached Karen, who knows a lot about poems, and asked \textit{``How was my poem?''}}}  & \textbf{Human}: You are talented. Put in more effort and it will be superb. & \textbf{Human:} I think your poem has merit and it's pretty good. \\  & \textbf{LLM:} I loved the effort you put into your poem and I think there's a lot of potential, but the rhythm and flow could use some improvement. & \textbf{LLM:} I liked most of it but there's definitely room for improvement in a few places. \\
\midrule
Imagine that John wanted to get Josh's opinion about a video game they just played. After Josh finished the game, John asked, \textit{``What did you think?''} & \textbf{Human:} I didn't really care for it, but I had fun hanging out with you. & \textbf{Human:} It was a really fun video game. \\
\bottomrule
\end{tabular}
\caption{Examples of open-ended human and LLM responses in Experiment 2.}
\label{tab:scenario-ratings}
\end{table*}

\section{Experiment 2 Methods} 
\label{sec:Exp2}

\subsection{Participants}

We recruited 156 participants through Prolific in the US or UK to take part in our open-ended response generation task.
Participants were compensated at a rate of \$15 / hour following the IRB protocol.

\subsubsection{Participant Demographics}

\begin{table}[t]
\centering

\renewcommand{\arraystretch}{1.2} 
\resizebox{\columnwidth}{!}{%
\begin{tabular}{lcccc}
\hline
\textbf{Stage} & \textbf{US} & \textbf{UK} & \textbf{Female} & \textbf{Male} \\
\hline
\begin{tabular}[c]{@{}l@{}}Open-ended Response \\ Collection (n=156)\end{tabular} 
& 80.1\% & 19.9\% & 59.7\% & 40.4\% \\
\begin{tabular}[c]{@{}l@{}}Human Evaluation \\ (n=156)\end{tabular}               
& 75.1\% & 24.9\% & 58.8\% & 40.7\% \\
\hline
\end{tabular}%
}
\caption{Participant Demographics}
\label{tab:participant_demographics}
\end{table}

See \autoref{tab:participant_demographics} for specific demographics of participants in our human studies.
To capture the diversity of our participant pool, we also collected self-reported ethnicity information. During the open-ended response collection stage, most participants identified as White (63.28\%), with the remainder identifying as Black (18.64\%), Mixed (7.91\%), Asian (5.08\%), or Other ethnic backgrounds (5.08\%). In the human evaluation stage, the distribution shifted slightly, with 57.31\% identifying as White, 18.71\% as Black, 10.53\% as Asian, 5.85\% as Mixed, and 7.02\% as Other ethnic backgrounds.

\subsection{Stimuli}

We first needed to elicit a large set of open-ended human responses to compare against the kinds of responses generated by LLMs.
To do this, we recruited $N=156$ participants through Prolific, located in the US or UK (compensated at a rate of
\$15/hour) and gave them an open textbox to imagine what someone would say in the given scenario.
Each participant was assigned 4 distinct scenarios out of the total set of 208 (see \ {fig:experiment-pipeline} middle panel). 
We planned our sample size to collect at least 3 different responses for each scenario.

To verify comprehension, we began with three warm-up questions featuring different ratings, requiring participants to simply match visual ratings with their textual equivalent. 
All participants effectively matched visuals with text, though five participants each made one error out of three questions. We still included their responses after manually reviewing them and confirming their alignment with the ratings and contexts presented.  
To minimize response bias and create a more naturalistic experience, we interspersed filler scenarios among the main testing scenarios. While structured identically to testing scenarios, filler scenarios focused on opinions about \emph{objects} rather than \emph{people} (see ~\autoref{tab:scenario-ratings} for examples).
Each participant thus viewed a total of 8 scenarios (4 main testing scenarios and 4 filler scenarios). We controlled the presentation to ensure that each participant was presented with a series of distinct stories, with each of the 4 goals and 4 true-state ratings appearing exactly once.

Next, we needed to collect responses from LLMs for comparison. 
Instead of the multiple-choice task we gave in the previous section,
Each model was presented with the same 208 scenarios as the human participants and was explicitly instructed to “keep your responses as short and concise as possible” to prevent excessively long answers. Each model
generated one response per scenario with a temperature setting of $\tau = 0$, resulting in a total of 624 responses collected.
We collected responses from three LLMs: \texttt{GPT-4o}, \texttt{Claude-3.5-Sonnet}, and \texttt{Llama-3.3-70B-Instruct}. 

\subsection{Design}

In the evaluation phase, we conducted a series of pairwise two-alternative forced choice comparisons, where human evaluators indicated which of a pair of responses they preferred for a given scenario.
We included three kinds of comparisons: 
\begin{enumerate} 
\item \textbf{\textit{Human vs. LLM preferences:}} Evaluators selected between human and LLM responses given identical scenarios, allowing us to understand which responses were preferred. 
\item \textbf{\textit{Goal Sensitivity:}} We compared responses generated for the original scenario (\textit{aligned-goal response}) against those generated for scenarios with different goals but identical ratings and contexts (\textit{misaligned-goal response}). This comparison revealed preferences between responses with aligned versus misaligned communicative goals.
\item \textbf{\textit{Rating Sensitivity:}} We presented pairs consisting of responses generated for the original scenario (\textit{aligned-rating response}) and responses generated with identical story and goal parameters but different ratings (\textit{misaligned-rating response}). This comparison identified preferences between responses with aligned versus misaligned ratings.
\end{enumerate}

\subsection{Procedure}
156 human evaluators are recruited from Prolific in the US or UK to take part in our evaluation task. Participants were compensated at a rate of \$15/ hour following the approved IRB protocol at <University Anonymized>.
Each participant evaluated four different scenarios with five preference questions per scenario: one trial comparing human vs. LLM responses, two trials assessing goal sensitivity within human and LLM sources, and two trials evaluating rating sensitivity for both sources.
We ensured that each participant was presented with responses from distinct human sources and distinct LLM sources in each block, and each participant completed a total of 4 blocks consisting of distinct scenarios with unique rating-goal combinations.
To minimize potential confounds, we implemented several additional controls. 
First, we randomized both question order and response option order within scenarios to control for order effects. 
We also inserted a transition page between blocks to reduce carryover effects. 
For the goal sensitivity and rating sensitivity comparisons, LLM comparisons were constrained to pairs of responses from the same model to control for model-specific variations in generation style.

\begin{figure*}[h!]
    \centering
    \includegraphics[width=1\textwidth]{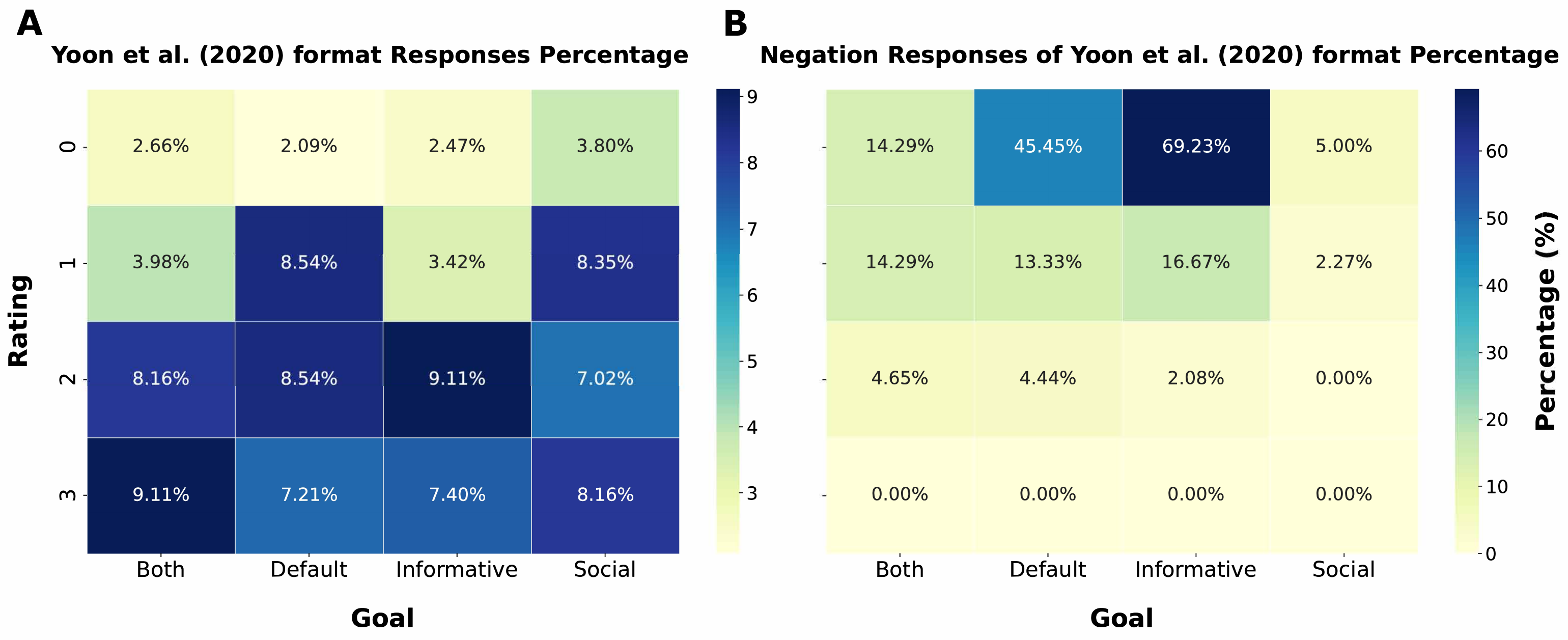} 
    \caption{(A) Distributions of how often the ``was/wasn't terrible/bad/good/amazing'' template studied by \citet{yoon20} was spontaneously produced by participants under each goal and rating. (B) How often responses use negation as a strategy among the responses that apply the \citet{yoon20} format under each goal and rating.}
    \label{fig:yoon-tessler analysis}
\end{figure*}

\subsection{Prompting strategy}

In the open-ended response case, we keep the whole scenario the same as in the multi-choice version, just omitting the answer options.

\textbf{Open-ended response generation system prompt: }\texttt{In the scenario, a person gave some performance and asked for another person's opinion on the performance. \texttt{\textbackslash n} The person's feelings in the scenario are shown on a scale of 0 to 3 hearts. \texttt{\textbackslash n} 0 heart, the lowest rating, means the person does not like the performance at all, and 3 hearts, the highest rating, means the person likes it a lot.\texttt{\textbackslash n} Please read the scenario carefully and answer the question in a complete sentence.\texttt{\textbackslash n} Please keep your responses as short and concise as possible!\texttt{\textbackslash n} Please only give the sentence-response without any other words!}

\subsection{Calibration on LLMs' open-ended generation}
\label{sec: calibartion}

To calibrate the choice of models in Experiment 2, we ran the open-ended experiment with \texttt{Llama-3.1-8B-Instruct}, one of the smaller models from Experiment 1, using an identical experimental setup. We collected 208 responses, one for each scenario. The results revealed severe limitations in the smaller model's capabilities. Despite receiving identical prompts (instructing concise responses), \texttt{Llama-3.1-8B} exhibited multiple failure modes: (1) it generated excessively long responses or incomplete sentences, (2) failed to follow basic instructions, and (3) frequently hallucinated content unrelated to the scenarios.

For example, when asked to provide informative feedback for a \emph{1-heart} rating, the model generated the following hallucinated ratings explanations:

\begin{quote}
    Don't try to be too different. Your presentation was good, but could have been smoother." Rating: 0 hearts (very direct). 
    
    This would not be true because Bob just wanted general feedback as opposed to negative feedback. Rating: 2 hearts (relatively kind but still true) (0, this might hurt Bob's feelings; 3, perfect)
\end{quote}

In more extreme cases, responses devolved into random character sequences after one or two coherent sentences:

\begin{quote}
 That was one of the least, uh...clear presentations I've heard in a while. However, I really like your confidence on stage, it's something you can work with! `` F A S L F B H M B F I S S M T B U S P K E Y T O I E N S P A S S I N G D E L I G H T E S S I G.''
\end{quote}

To enable a meaningful comparison with models used in the main analysis, we truncated all \texttt{Llama-3.1-8B} responses to the first complete sentence and annotated them for politeness strategies using the same methodology as our main analysis. The resulting distributions differed significantly from all larger models tested, with Jensen-Shannon Divergence values exceeding 0.53 ($p < 0.001$) in all comparisons. These findings suggest that \emph{model size} represents a critical threshold for pragmatic competence, even if we ignore basic failures to follow task instructions. While we cannot isolate whether this stems from training data, reinforcement strategies, or architectural limitations, this calibration experiment confirms our initial decision to focus on models that demonstrate basic pragmatic competence. Additionally, the investigation of size thresholds will be an important direction of future work.

\section{Additional details of linguistic analysis}
\label{sec:annotation}

\begin{table}[t]
\centering
\resizebox{\columnwidth}{!}{%
\begin{tabular}{lcc}
\toprule
\textbf{Groups} & \textbf{Observed JSD} & \textbf{Null Means}\\
\midrule
preferred vs. non-preferred  & 0.013 &0.009\\
human vs. LLM  & 0.175 & 0.058 \\
both vs. informative  & 0.134 & 0.081\\
both vs. social & 0.195 & 0.096\\
both vs. default & 0.136 & 0.088\\
informative vs. social & 0.231 & 0.100\\
informative vs. default & 0.134 & 0.093 \\
social vs. default & 0.166 & 0.106 \\
0 hearts vs. 1 heart & 0.119 & 0.087\\
0 hearts  vs. 2 hearts & 0.167 & 0.089 \\
0 hearts  vs. 3 hearts & 0.227 & 0.094\\
1 heart  vs. 2 hearts & 0.129 & 0.088 \\
1 heart vs. 3 hearts & 0.225 & 0.093\\
2 hearts  vs. 3 hearts & 0.191 & 0.095\\
\bottomrule
\end{tabular}
}
\caption{JSD with word frequency counting distribution, all the p-values are \textit{< .001}}
\label{tab:jsd-word-frequency}
\end{table}

\begin{table}[t]
\centering

\resizebox{\columnwidth}{!}{%
\begin{tabular}{lcc}
\toprule
\textbf{Groups} & \textbf{Observed JSD} & \textbf{Null Means}\\
\midrule
gpt4.1 vs. claude3.7 labels  & 0.045 & 0.004\\
gpt4.1 vs. golden labels  & 0.073 & 0.004\\
claude3.7 vs. golden labels  & 0.026 & 0.003\\
preferred vs. non-preferred golden labels  & 0.008 (p = 0.087) & 0.006 \\
human vs. LLM response golden labels  & 0.023 & 0.006 \\
both vs. informative  & 0.060 & 0.011\\
both vs. social & 0.107 & 0.011\\
both vs. default & 0.033 & 0.011\\
informative vs. social & 0.180 & 0.013\\
informative vs. default & 0.0197 (p = 0.006) & 0.0125 \\
social vs. default & 0.131 & 0.013 \\
0 hearts vs. 1 heart & 0.049 & 0.011\\
0 hearts  vs. 2 hearts & 0.115 & 0.012 \\
0 hearts  vs. 3 hearts & 0.286 & 0.013\\
1 heart  vs. 2 hearts & 0.079 & 0.011 \\
1 heart vs. 3 hearts & 0.263 & 0.012\\
2 hearts  vs. 3 hearts & 0.115 & 0.012\\
\bottomrule
\end{tabular}
}
\caption{JSD with politeness strategy frequency counting distribution, all the p-values are \textit{< .001} unless specified. We checked the agreement between \texttt{gpt-4.1, claude-3.7-sonnet}, and golden labels and found out the differences are quite significant (p < .001 - non-trivial). We compared the golden-labeled politeness strategies between human and LLM responses; we all use the golden-labeled politeness strategies for comparisons between goals and ratings. }
\label{tab:jsd-politeness-strategy}
\end{table}

\subsection{JSD tables}

See the complete JSD scores of word-frequency distribution and politeness-strategy distribution at \autoref{tab:jsd-word-frequency} and~\ref{tab:jsd-politeness-strategy}.

\subsection{Text classification with SBERT embeddings}
\label{sec:text-classifiers}

To analyze if there are high-dimensional features that differentiate each group beyond simple statistical analysis, we trained several simple classifiers using SBERT \cite{sbert-2019}, applying the pretrained sentence transformer ``all-MiniLM-L6-v2'' to generate the response embeddings. Four different classifiers were implemented: logistic regression, random forest, SVM, and a simple MLP with a hidden layer size of 100 using \texttt{scikit-learn}. We then analyzed the results using the best-performing model among these four approaches.

Our analysis revealed that predicting between preferred and non-preferred responses is challenging, with performance only slightly above chance at approximately 54\% across F1 score, recall, and precision metrics. In contrast, identifying human versus LLM responses was significantly more predictable, with the classifier reaching about 83\% accuracy across the same metrics.

When examining the four goals comparison, we observed varying levels of predictability. Considering all responses collectively (both human and LLM), the classifier achieved approximately 60\% performance in distinguishing between the four goals. Interestingly, when analyzing LLM responses in isolation, predictability increased to approximately 70\%. whereas focusing solely on human responses, predictability decreased to around 40\%. This suggests that LLM responses contain more distinctive patterns associated with different communicative goals compared to human responses, which exhibit greater variability in their approach to achieving the same goals.




\subsection{Politeness annotation process}
\label{sec:politeness-annotation}

To handle the cases where a single politeness marker can be categorized under different politeness strategies, we allow the LLMs to assign up to three strategies per marker. Given our observation that both humans and LLMs often mix different politeness strategies in a single response, we also instruct the LLMs to identify all markers they consider reasonable. In the system prompt, we provided a comprehensive list of politeness strategies mainly from  \citet{brown1987politeness} framework with additional ones from \citet{danescu-niculescu-mizil-etal-2013-computational}. (see Appendix~\ref{sec:llm-annotation-prompt}). By providing a predefined, finite list of politeness strategies, we hope to unify the distribution of politeness strategies that these two LLMs can choose from and make their annotation results comparable. Since LLMs can make mistakes and often hallucinate \cite{xu2025hallucinationinevitableinnatelimitation, Huang_2025} We manually inspected every single one of the labels annotated by the two LLMs, observing differences between the annotation results, and re-annotated any that we thought were not correctly labeled by the LLMs based on the definitions and examples provided in the two frameworks as the golden labels.

There are still many cases where different researchers can label differently; our golden labels are just an instantiation of how we think what politeness strategies should be attached to politeness markers. 
We also note that ``negation'' is not considered as a specific politeness strategy in these two works, and thus we do not include it as a politeness strategy in the provided comprehensive list; we consider it as a ``positive politeness - avoid disagreement'' or ``off-record - understate'' when it appears based on the contexts. 
Throughout manual inspection, we indeed found out LLMs sometimes are not consistent with their labels - the same words can be labeled differently in different responses under the same goal and rating, and they sometimes hallucinate and give politeness strategies that are not in the provided list.

For the comprehensive list of politeness strategies provided in the annotation system prompt, there are several things to notice. First, the list is a combination of \citet{brown1987politeness} and \citet{danescu-niculescu-mizil-etal-2013-computational}. We chose the list of politeness strategies from \citet{brown1987politeness} because their classic framework covers nearly all politeness strategies and is still widely used and adopted in most current work on politeness. The list of politeness strategies shown in \citet{danescu-niculescu-mizil-etal-2013-computational} is mainly adopted from \citet{brown1987politeness}, with some additional strategies based on some other widely used politeness phenomena and literature. We believe that by combining the two lists, we can obtain a comprehensive set of all widely used politeness strategies in language.

A note on combining the two lists: if there is any disagreement between the two works, we follow the categorization in \citet{brown1987politeness}. Specifically, we consider Deference a negative politeness strategy, in line with \citet{brown1987politeness}, whereas in \citet{danescu-niculescu-mizil-etal-2013-computational}, it is categorized as a positive strategy.

Our manual inspection of \texttt{gpt-4o} and \texttt{claude-3.7-sonnet} and golden-label generation follows several principles: 

\begin{itemize}
    \item We follow the definitions of politeness strategies provided in their respective frameworks and annotate all politeness markers in a given response. A single politeness marker may correspond to multiple politeness strategies, and people often mix different strategies within a single response. We consider all politeness markers and label each one with up to three of the most significant politeness strategies.
    \item For words that are not clearly significant enough to be considered politeness markers—cases where they could be interpreted as either common words or politeness markers—we simply accept whatever the LLMs produce. 
    \item Could/Would are counterfactual modals, which are widely used in polite speech. They are not considered in \citet{brown1987politeness}, but are included in \citet{danescu-niculescu-mizil-etal-2013-computational}. In our manual labeling process, we always mark them as counterfactual modals and additionally label them with other relevant politeness strategies, such as hedging, when appropriate.
\end{itemize}

\subsection{LLM annotation prompt}
\label{sec:llm-annotation-prompt}

\textbf{The following whole section is the complete system prompt:}

{\ttfamily
\noindent You are an expert in the study of human polite language use, with extensive knowledge of the relevant literature and the various politeness strategies people employ in everyday conversation. 

\noindent Please follow my instructions to help extract, annotate, and categorize politeness markers used in the response of each scenario.

\noindent In each scenario, Person A asks Person B for their opinion on A's performance. Person B's true feelings are represented on a scale of 0 to 3 hearts as the rating, where 0 hearts means they did not like the performance at all, and 3 hearts means they liked it very much. 

\noindent In the question, please pay attention to the communicative goal mentioned (either to be informative, to make person A feel good, to do both, or to serve as the default with no specific goal).

\noindent Your tasks are to:
\begin{enumerate}
\item Read the whole scenario setup carefully, pay attention to the rating and the communicative goal in the question. Then, identify and annotate the specific word(s) or phrase(s) in Person B's response that function as politeness markers.
\item Categorize each politeness marker using the comprehensive list of politeness strategies provided below, specifying both the category (positive politeness, negative politeness, off-record, bald-on-record) and the corresponding specific politeness strategy.
\end{enumerate}

\noindent Please present your answer in the following format for each politeness marker WITHOUT ANY additional text or explanation:

\begin{quote}
Politeness marker: [the specific word(s) or phrase(s)]\\
Politeness strategy-1: [category + specific politeness strategy] \\
Politeness strategy-2: [category + specific politeness strategy] (if applicable)\\
Politeness strategy-3: [category + specific politeness strategy] (if applicable)\\

(Repeat the above for each politeness marker found in the response)
\end{quote}

\noindent Below is the comprehensive list of politeness strategies with examples for each strategy. The politeness markers, e.g., the specific word(s) or phrase(s) used in each strategy's example, are shown in parentheses.

\noindent Please pay attention to the usage of could/would or similar words in the following list.

\bigskip

\noindent\textbf{I. Positive Politeness Strategies}
\begin{enumerate}

\item \textbf{Gratitude}
\begin{itemize}
\item Example 1: ``Thank you so much for your help!'' (thank you)
\item Example 2: ``I really appreciate your kindness.'' (I really appreciate)
\end{itemize}

\item \textbf{Greeting (social approach)}
\begin{itemize}
\item Example 1: ``Hi there! Could you help me out?'' (Hi there)
\item Example 2: ``Good morning! How are you today?'' (Good morning)
\end{itemize}

\item \textbf{Greeting (social approach)}
\begin{itemize}
\item Example 1: ``Hi there! Could you help me out?'' (Hi there)
\item Example 2: ``Good morning! How are you today?'' (Good morning)
\end{itemize}

\item \textbf{Positive Lexicon (positive sentiment, optimism)}
\begin{itemize}
\item Example 1: ``Wow, that's wonderful news!'' (wonderful)
\item Example 2: ``I'm thrilled about your promotion.'' (thrilled)
\end{itemize}

\item \textbf{Notice, attend to hearer's interests, wants, needs}
\begin{itemize}
\item Example 1: ``You seem stressed—can I assist?'' (You seem stressed)
\item Example 2: ``You must be tired, please take a rest.'' (You must be tired)
\end{itemize}

\item \textbf{Exaggerate interest, approval, sympathy}
\begin{itemize}
\item Example 1: ``That's the best presentation I've ever seen!'' (the best)
\item Example 2: ``Your idea is absolutely fantastic!'' (absolutely fantastic)
\end{itemize}

\item \textbf{Intensify interest in hearer}
\begin{itemize}
\item Example 1: ``I traveled across town just to see you!'' (just to see you)
\item Example 2: ``I've been eagerly waiting to hear your story.'' (eagerly waiting)
\end{itemize}

\item \textbf{Use in-group identity markers}
\begin{itemize}
\item Example 1: ``Hey mate, can you give me a hand?'' (mate)
\item Example 2: ``Buddy, I need your advice on something.'' (Buddy)
\end{itemize}

\item \textbf{Seek agreement}
\begin{itemize}
\item Example 1: ``It's beautiful today, isn't it?'' (isn't it?)
\item Example 2: ``This solution seems ideal, right?'' (right?)
\end{itemize}

\item \textbf{Avoid disagreement}
\begin{itemize}
\item Example 1: ``Yes, that might work, but also consider…'' (that might work)
\item Example 2: ``I see your point, though perhaps…'' (I see your point)
\end{itemize}

\item \textbf{Presuppose/assert common ground}
\begin{itemize}
\item Example 1: ``You know how much we both value honesty.'' (we both)
\item Example 2: ``We both know how difficult this can be.'' (We both know)
\end{itemize}

\item \textbf{Joke}
\begin{itemize}
\item Example 1: ``If you fix this bug, I'll bake you cookies!'' (I'll bake you cookies)
\item Example 2: ``Careful, your brilliance is showing!'' (your brilliance is showing)
\end{itemize}

\item \textbf{Assert speaker's knowledge of hearer's wants}
\begin{itemize}
\item Example 1: ``Since I know you like chocolate, here's a cake.'' (Since I know you like chocolate)
\item Example 2: ``Knowing you love adventure, I booked a trip.'' (Knowing you love adventure)
\end{itemize}

\item \textbf{Offer, promise}
\begin{itemize}
\item Example 1: ``I'll take care of that for you tomorrow.'' (I'll take care)
\item Example 2: ``If you're busy, I promise to handle it myself.'' (I promise)
\end{itemize}

\item \textbf{Be optimistic}
\begin{itemize}
\item Example 1: ``I'm sure you can easily solve this.'' (I'm sure)
\item Example 2: ``You'll definitely manage to finish this in time.'' (You'll definitely)
\end{itemize}

\item \textbf{Include both speaker and hearer (inclusive 'we')}
\begin{itemize}
\item Example 1: ``Let's figure this out together.'' (Let's)
\item Example 2: ``Why don't we start the project now?'' (we)
\end{itemize}

\item \textbf{Give or ask for reasons}
\begin{itemize}
\item Example 1: ``Could you come with me? It'll be helpful.'' (It'll be helpful)
\item Example 2: ``Why not join the group? You'd enjoy it.'' (You'd enjoy it)
\end{itemize}

\item \textbf{Assume reciprocity}
\begin{itemize}
\item Example 1: ``You helped me last time, now it's my turn.'' (now it's my turn)
\item Example 2: ``I lent you my notes earlier—can I borrow yours today?'' (I lent you my notes earlier)
\end{itemize}

\item \textbf{Give gifts to hearer (sympathy, understanding, cooperation)}
\begin{itemize}
\item Example 1: ``You've been working hard; here's a small gift.'' (here's a small gift)
\item Example 2: ``Here, take this coffee—you deserve a break.'' (you deserve a break)
\end{itemize}
\end{enumerate}

\noindent\textbf{II. Negative Politeness Strategies}
\begin{enumerate}
\item \textbf{Apologizing}
\begin{itemize}
\item Example 1: ``Sorry to disturb you, but I have a question.'' (Sorry)
\item Example 2: ``I apologize for interrupting your meeting.'' (I apologize)
\end{itemize}

\item \textbf{Please (sentence-medial polite form)}
\begin{itemize}
\item Example 1: ``Could you please send me the document?'' (please)
\item Example 2: ``Would you please consider my suggestion?'' (please)
\end{itemize}

\item \textbf{Be conventionally indirect}
\begin{itemize}
\item Example 1: ``Could you possibly close the door?'' (Could you possibly)
\item Example 2: ``Would you mind handing me the pen?'' (Would you mind)
\item Example 3: ``By the way, do you know the time?'' (By the way)
\item Example 4: ``Oh, by the way, did you finish the report?'' (Oh, by the way)
\end{itemize}

\item \textbf{Question, hedge}
\begin{itemize}
\item Example 1: ``Perhaps we could reconsider the deadline?'' (Perhaps)
\item Example 2: ``Maybe you might find this helpful?'' (Maybe)
\item Example 3: ``I suggest we might consider other options.'' (might consider)
\item Example 4: ``I think it's possibly better this way.'' (I think, possibly)
\end{itemize}

\item \textbf{Be pessimistic}
\begin{itemize}
\item Example 1: ``I don't suppose you could spare a moment?'' (I don't suppose)
\item Example 2: ``You probably wouldn't want to help, would you?'' (probably wouldn't want)
\end{itemize}

\item \textbf{Minimize the imposition}
\begin{itemize}
\item Example 1: ``I just need a quick moment of your time.'' (just need a quick moment)
\item Example 2: ``This will take only a second, I promise.'' (only a second)
\end{itemize}

\item \textbf{Give deference}
\begin{itemize}
\item Example 1: ``Professor, could you clarify this point?'' (Professor)
\item Example 2: ``Excuse me, sir, may I interrupt?'' (Excuse me, sir)
\end{itemize}

\item \textbf{Impersonalize speaker and hearer}
\begin{itemize}
\item Example 1: ``It seems this task needs attention.'' (It seems)
\item Example 2: ``There appears to be a misunderstanding.'' (There appears)
\end{itemize}

\item \textbf{State the FTA as a general rule}
\begin{itemize}
\item Example 1: ``Visitors are requested not to use cell phones.'' (Visitors are requested)
\item Example 2: ``Eating is not allowed in the library.'' (is not allowed)
\end{itemize}

\item \textbf{Nominalize}
\begin{itemize}
\item Example 1: ``Your participation is required.'' (participation)
\item Example 2: ``Submission of your paper is expected soon.'' (Submission)
\end{itemize}

\item \textbf{Go on record incurring a debt}
\begin{itemize}
\item Example 1: ``I'd greatly appreciate it if you helped me.'' (I'd greatly appreciate it)
\item Example 2: ``I'll owe you one if you can cover my shift.'' (I'll owe you one)
\end{itemize}

\item \textbf{Counterfactual modal forms (could/would)}
\begin{itemize}
\item Example 1: ``Could you assist me with this?'' (Could you)
\item Example 2: ``Would you mind checking this for me?'' (Would you mind)
\end{itemize}

\item \textbf{Indicative modal forms (can/will)}
\begin{itemize}
\item Example 1: ``Can you help me with these files?'' (Can you)
\item Example 2: ``Will you be able to come by later?'' (Will you)
\end{itemize}
\end{enumerate}

\noindent\textbf{III. Off-Record (Indirect) Strategies}
\begin{enumerate}
\item \textbf{Give hints}
\begin{itemize}
\item Example 1: ``It's chilly in here…'' (chilly in here) – hint to close the window
\item Example 2: ``I'm thirsty.'' (I'm thirsty) – hint to offer a drink
\end{itemize}

\item \textbf{Give association clues}
\begin{itemize}
\item Example 1: ``Oh no, I forgot my wallet!'' (forgot my wallet) – hint to pay for them
\item Example 2: ``My phone just died.'' (phone just died) – hint to borrow a phone
\end{itemize}

\item \textbf{Presuppose}
\begin{itemize}
\item Example 1: ``I cleaned it again today.'' (again) – presupposes someone else didn't
\item Example 2: ``Did you check the oven?'' (Did you check) – implies concern or oversight
\end{itemize}

\item \textbf{Understate}
\begin{itemize}
\item Example 1: ``The movie was not exactly thrilling.'' (not exactly thrilling)
\item Example 2: ``His speech was somewhat unclear.'' (somewhat unclear)
\end{itemize}

\item \textbf{Overstate}
\begin{itemize}
\item Example 1: ``I've waited forever for your reply!'' (waited forever)
\item Example 2: ``I'm starving!'' (starving)
\end{itemize}

\item \textbf{Tautologies}
\begin{itemize}
\item Example 1: ``Business is business.'' (Business is business)
\item Example 2: ``It is what it is.'' (It is what it is)
\end{itemize}

\item \textbf{Contradictions}
\begin{itemize}
\item Example 1: ``It's good, but at the same time, not good.'' (good, but not good)
\item Example 2: ``I'm happy and not happy about this.'' (happy and not happy)
\end{itemize}

\item \textbf{Be ironic}
\begin{itemize}
\item Example 1: ``Lovely day we're having!'' (Lovely day) – during bad weather
\item Example 2: ``That went well!'' (That went well) – after a failure
\end{itemize}

\item \textbf{Use metaphors}
\begin{itemize}
\item Example 1: ``He's got a heart of stone.'' (heart of stone)
\item Example 2: ``She's a ray of sunshine.'' (ray of sunshine)
\end{itemize}

\item \textbf{Rhetorical questions}
\begin{itemize}
\item Example 1: ``How many times must I tell you?'' (How many times)
\item Example 2: ``Do I look like I'm joking?'' (Do I look)
\end{itemize}

\item \textbf{Be ambiguous}
\begin{itemize}
\item Example 1: ``Something feels off about this.'' (feels off)
\item Example 2: ``It seems unusual somehow…'' (seems unusual)
\end{itemize}

\item \textbf{Be vague}
\begin{itemize}
\item Example 1: ``I'm a bit upset.'' (a bit)
\item Example 2: ``I kind of disagree.'' (kind of)
\end{itemize}

\item \textbf{Over-generalize}
\begin{itemize}
\item Example 1: ``Everyone knows it's not true.'' (Everyone knows)
\item Example 2: ``Nobody likes that.'' (Nobody)
\end{itemize}

\item \textbf{Displace hearer}
\begin{itemize}
\item Example 1: ``I wish someone would help.'' (someone)
\item Example 2: ``It'd be great if someone cleaned up.'' (someone)
\end{itemize}

\item \textbf{Be incomplete (ellipsis)}
\begin{itemize}
\item Example 1: ``If only you knew…'' (If only you knew)
\item Example 2: ``Well, if you could just…'' (if you could just)
\end{itemize}
\end{enumerate}

\noindent\textbf{IV. Bald-on-Record Strategies}
\begin{enumerate}
\item \textbf{Direct questions/statements}
\begin{itemize}
\item Example 1: ``What are you doing?'' (What are you doing?)
\item Example 2: ``Where did you put it?'' (Where did you put it?)
\end{itemize}

\item \textbf{Direct commands (imperatives)}
\begin{itemize}
\item Example 1: ``Stop right now!'' (Stop)
\item Example 2: ``Bring it to me immediately.'' (Bring it)
\end{itemize}

\item \textbf{Sentence-initial imperative forms (``Please'' start—less polite)}
\begin{itemize}
\item Example 1: ``Please move out of my way.'' (Please move)
\item Example 2: ``Please finish your work quickly.'' (Please finish)
\end{itemize}

\item \textbf{Sentence-initial second-person statements (less polite)}
\begin{itemize}
\item Example 1: ``You need to fix this.'' (You need to)
\item Example 2: ``You've misunderstood me.'' (You've misunderstood)
\end{itemize}

\item \textbf{Factuality (direct assertions, less polite)}
\begin{itemize}
\item Example 1: ``Actually, you did it incorrectly.'' (you did it incorrectly)
\item Example 2: ``The truth is you failed to deliver.'' (you failed to deliver)
\end{itemize}

\item \textbf{Negative lexicon (negative sentiment, impolite)}
\begin{itemize}
\item Example 1: ``You're always messing things up!'' (always messing things up)
\item Example 2: ``If you're going to accuse me…'' (accuse me)
\end{itemize}
\end{enumerate}

}

\subsection{Guidelines for correcting LLM annotation results}

We calculated the initial agreement between our two LLM annotators (\texttt{GPT-4.1} and \texttt{Claude-3.7-Sonnet}) at 39.8\% (using a strict measure requiring an exact match of the set of strategies), indicating substantial divergence in their initial classifications. When compared to the human-corrected “gold standard” labels, \texttt{Claude-3.7-Sonnet} achieved 34.3\% agreement while \texttt{GPT-4.1} achieved 36.1\% agreement. Because each utterance could receive multiple strategy labels, we observed that many LLM annotations contained a mix of appropriate and inappropriate labels. Rather than rejecting these labels entirely, we systematically corrected them by removing inappropriate labels and retaining valid ones. The authors first established the correction criteria through discussion, then independently reviewed annotations.

We identified four main categories of LLM annotation errors:

\begin{enumerate}
    \item LLM annotators often labelled strategies without considering the scenario context that we provided. For instance, ``there is still room for improvement'' was labeled as an ``understatement'' regardless of the true rating. This label would be appropriate for poor ratings (0/3 hearts) but inappropriate when the true rating was already positive (2/3 hearts).
    \item Relatedly, when ratings literally matched the content (e.g., ``really good'' for a 3-heart rating), LLMs annotators sometimes incorrectly tagged them as ``exaggeration'' strategies.
    \item In about 20 cases, LLMs generated strategies not present in the prompt, such as ``Negative Politeness – Be specific'', which doesn't exist in the taxonomy of \citet{brown1987politeness}.
    \item In a remaining 5-10\% of cases, annotations were entirely wrong, such as labeling ``that really makes a solo shine'' as a hedge when it clearly expresses a positive evaluation.
\end{enumerate}



\subsection{A comprehensive list of politeness strategies with examples}

See Table~\ref{tab:politeness-strategies-list} for a comprehensive list of politeness strategies used in both human and LLM responses. The examples and strategies shown are based on golden labels from our collected responses.

\begin{table*}[ht]
\centering
\setlength{\tabcolsep}{4pt} 
\renewcommand{\arraystretch}{0.88} 
\begin{tabularx}{\textwidth}{l lX}
\toprule
\textbf{Category} & \textbf{Specific Politeness Strategy} & \multicolumn{1}{c}{\textbf{Example}} \\
\midrule
\multirow{16}{*}{\parbox{2cm}{Positive\\Politeness}} 
& 1. Assert speaker's knowledge of hearer's wants & \textit{I know }you are up to the challenge! \\
& 2. Avoid disagreement & \textit{Pretty decent} for a beginner\\
& 3. Be optimistic & You can even make them better next time!\\
& 4. Exaggerate interest, approval, sympathy & Your dance \textit{greatly exceeded all }expectations.\\
& 5. Give gifts to hearer & I am so proud of you.\\
& 6. Give or ask for reasons & I have tasted some really good cakes, and yours ...\\
& 7. Gratitude & I'm so grateful that \ldots \\
& 8. Greeting & \textit{Hey}, I read your review \ldots \\
& 9. Include both speaker and hearer & \textit{Let's} go through it together \\
& 10. Intensify interest in hearer & You were born to be on stage\\
& 11. Notice, attend to hearer's interests & I can see you put in lots of effort\\
& 12. Offer, promise & Let me know if you need any tips. \\
& 13. Positive lexicon & It was absolutely amazing! \\
& 14. Presuppose/assert common ground & \ldots with other artists of your caliber \\
& 15. Seek agreement & Is this your first time baking?\\
& 16. Use in-group identity markers & Your app is pretty good, \textit{Henry}!\\

\midrule
\multirow{8}{*}{\parbox{2cm}{Negative\\Politeness}} 
& 17. Apologizing & I didn't like it, sorry!\\
& 18. Be conventionally indirect & If you would like \ldots\\
& 19. Counterfactual modal forms & Could/Would you \ldots\\
& 20. Give deference & \textit{In my expert opinion,} your painting is terrible. \\
& 21. Impersonalize speaker and hearer & \textit{There are} a few places to improve.\\
& 22. Minimize the imposition & I have \textit{a few} suggestions (0 +social)  \\
& 23. Nominalize & I would not be the best person to evaluate \textit{your performance}. \\
& 24. Question, hedge & Maybe try adding some different flavoring ingredients next time?\\
\midrule
\multirow{10}{*}{\parbox{2cm}{Off-Record}} 
& 25. Be ironic & It was horrible, my eyes are bleeding.\\
& 26. Be vague & It was interesting (0-rating case)\\
& 27. Contradictions & It was great, however, it needs improvement\\
& 28. Displace hearer & You looked so confident and elegant! (when commenting on performance)\\
& 29. Give association clues & Better than \textit{those who can't play} \\
& 30. Give hints & It could be better if you adjusted the sweetness!\\
& 31. Overstate & The cookies tasted great. (1+social)\\
& 32. Presuppose & Pretty decent \textit{for a beginner}\\
& 33. Understate & It was not good (0-hearts rating)\\
& 34. Use metaphors & Your singing was like music to my ears!\\
\midrule
\multirow{5}{*}{\parbox{2cm}{Bald on-\\Record}} 
& 35. Direct commands & Try practicing with precise measurements.\\
& 36. Factuality & I didn't like the cookies at all.\\
& 37. Negative lexicon & It was terrible!\\
& 38. Sentence-initial 2nd-person statements & You need to work on that.\\
& 39. Sentence-initial imperative forms & Please for gods sake improve on these areas \\
\bottomrule
\end{tabularx}
\caption{A comprehensive list of politeness strategies with examples from our collected responses. We consider all the politeness strategies and politeness markers in the golden annotation results.}
\label{tab:politeness-strategies-list}
\end{table*}

\end{document}